\documentclass[twoside,11pt]{article}

\usepackage{algorithm}
\usepackage{algorithmic}
\usepackage{times}
\usepackage{graphicx} % more modern
\usepackage{subfigure} 
\usepackage{caption}
\usepackage{natbib}
\usepackage{amsmath}
\usepackage{color}
\usepackage{jmlr}

\newcommand\blfootnote[1]{%
  \begingroup
  \renewcommand\thefootnote{}\footnote{#1}%
  \addtocounter{footnote}{-1}%
  \endgroup
}
\ShortHeadings{Graying the black box: Understanding DQNs}{Zahavy, Baram and Mannor}
\firstpageno{1}

\begin{document}

\title{Graying the black box: Understanding DQNs}
\author{\name Tom Zahavy* \email tomzahavy@campus.technion.ac.il 
       \AND
       \name Nir Baram* \email nirb@tx.technion.ac.il
       \AND
       \name Shie Mannor \email shie@ee.technion.ac.il\\\\
       Department of Electrical Engineering\\
       The Technion, Israel Institute of Technology \\
       Haifa, 32000, Israel\\\\
       *These authors have contributed equally.
}

\editor{}    
\maketitle
    
\begin{abstract}

In recent years there is a growing interest in using deep representations for reinforcement learning. In this paper, we present a methodology and tools to analyze Deep Q-networks (DQNs) in a non-blind matter. Moreover, we propose a new model, the Semi Aggregated Markov Decision Process (SAMDP), and an algorithm that learns it automatically. The SAMDP model allows us to identify spatio-temporal abstractions directly from features and may be used as a sub-goal detector in future work. Using our tools we reveal that the features learned by DQNs aggregate the state space in a hierarchical fashion, explaining its success. Moreover, we are able to understand and describe the policies learned by DQNs for three different Atari2600 games and suggest ways to interpret, debug and optimize deep neural networks in reinforcement learning.

\end{abstract}
    
\section{Introduction}
\blfootnote{This paper was done with the support of the Intel Collaborative Research institute for Computational Intelligence (ICRI-CI) and is part of the 'Why $\&$ When Deep Learning works $–$ looking inside Deep Learning' ICRI-CI paper bundle. Some of the results in this paper were originally published in \citep{Zahavy2016,visdynamics,visdynamics2}}

In the Reinforcement Learning (RL) paradigm, an agent autonomously learns from experience in order to maximize some reward signal. Learning to control agents directly from high-dimensional inputs like vision and speech is a long-standing problem in RL, known as the curse of dimensionality. Countless solutions to this problem have been offered including linear function approximators \citep{tsitsiklis1997analysis}, hierarchical representations \citep{dayan}, state aggregation \citep{singh1} and options \citep{sutton1999between}. These methods rely upon engineering problem-specific state representations, hence, reducing the agent's flexibility and making the learning more tedious. Therefore, there is a growing interest in using nonlinear function approximators, that are more general and require less domain specific knowledge, e.g., TD-gammon \citep{tesauro1995temporal}. Unfortunately, such methods are known to be unstable or even to diverge when used to represent the action-value function \citep{tsitsiklis1997analysis,gordon1995stable,riedmiller2005neural}.

The Deep Q-Network (DQN) algorithm \citep{mnih2015human} increased training stability by introducing the target network and by using Experience Replay (ER) \citep{lin1993reinforcement}. Its success was demonstrated in the Arcade Learning Environment (ALE) \citep{bellemare2012arcade}, a challenging framework composed of dozens of Atari games used to evaluate general competency in AI. DQN achieved dramatically better results than earlier approaches, showing a robust ability to learn representations.\\

While using Deep Learning (DL) in RL seems promising, there is no free lunch. Training a Deep Neural Network (DNNs) is a complex two-levels optimization process. At the inner level, we fix a hypothesis class and learn it using some gradient descent method. The optimization of the outer level addresses the choice of network architecture, setting hyper-parameters, and even the choice of the optimization algorithm. While the optimization of the inner level is analytic to a high degree, the optimization of the outer level is far from being so. Currently, most practitioners tackle this problem by either a trial and error methodology, or by exhaustively searching over the space of possible configurations. Moreover, in Deep Reinforcement Learning (DRL) we also need to choose how to model the environment as a Markov Decision Process (MDP), i.e., to choose the discount factor, the amount of history frames that represent a state, and to choose between the various algorithms and architectures \citep{nair2015massively,van2015deep,schaul2015prioritized,wang2015dueling,bellemare2015increasing}.\\

A key issue in RL is that of representation, i.e., allowing the agent to locally generalize the policy across similar states. Unfortunately, spatially similar states often induce different control rules ("labels") in contrast to other machine learning problems that enjoy local generalization. In fact, in many cases the optimal policy is not a smooth function of the state, thus using a linear approximation is far from being optimal. For example, in the Atari2600 game Seaquest, whether or not a diver has been collected (represented by a few pixels) controls the outcome of the submarine surface action (with: fill air, without: loose a life) and therefore the optimal policy. Another cause for control discontinuities is that for a given problem, two states with similar representations may, in fact, be far from each other in terms of the number of state transitions required to reach one from the other. This observation can also explain the lack of pooling layers in DRL architectures \citep{mnih2015human,mnih2013playing,levine2015end}. \\

Different methods that focus on the temporal structure of the policy, has also been proposed. Such methods decompose the learning task into simpler subtasks using graph partitioning \citep{menache2002q,mannor2004dynamic,csimcsek2005identifying} and path processing mechanisms \citep{stolle2004automated, thrun1998learning}. Given a good temporal representation of the states, varying levels of convergence guarantees can be promised \citep{dean1995decomposition,parr1998flexible,hauskrecht1998hierarchical,dietterich2000hierarchical}.\\

Our main observation in this work is that DQN is learning an internal model \citep{francis1975internal,sontag2003adaptation} of the domain without explicitly being trained to. We discover the internal model by identifying spatio-temporal abstractions directly from the learned representation in two different ways: (1) manually clustering the state space using hand crafted features, and (2), the Semi Aggregated Markov Decision Process (SAMDP), an approximation of the true MDP that is learned automatically from data. We show that by doing so we can offer an interpretation of learned policies and discover the policy hierarchy.\\
In particular, our main contributions are:

\begin{itemize}
\item \textbf{Understanding:} We show that DQNs are learning temporal abstractions of the state space such as hierarchical state aggregation and options. Temporal abstractions were known to the RL community before as mostly manual tools to tackle the curse of dimensionality; however, we observe that a DQN is finding abstractions automatically. Thus, we believe that our analysis explains the success of DRL from a reinforcement learning research perspective.
\item \textbf{Interpretability:} We give an interpretation of the agent's policy in a clear way, thus allowing to understand what are its weaknesses and strengths.
\item \textbf{Debugging:} We propose debugging methods that help to reduce the hyper parameters grid search of DRL. Examples are given on game modeling, termination and initial states and score over-fitting.
\end{itemize}

\section{Background and Related Work}
\subsection{RL}
The goal of RL agents is to maximize its expected total reward by learning an optimal policy (mapping  states to actions). At time $t$ the agent observes a state $s_t$, selects an action $a_t$, and receives a reward $r_t$, following the agents decision it observes the next state $s_{t+1}$ . We consider infinite horizon problems where the cumulative return is discounted by a factor of $\gamma\in[0,1]$ and the return at time $t$ is given by $R_t = \sum_{t'=t}^T\gamma^{t'-t}r_t$, where T is the termination step. The action-value function $Q^{\pi}(s,a)$ measures the expected return when choosing action $a_t$ at state $s_t$, and following policy $\pi$: $Q(s,a) = \mathbb{E} [R_t|s_t = s, a_t = a, \pi]$ afterwards. The optimal action-value obeys a fundamental recursion known as the Bellman equation, \begin{equation*}
Q^* (s_t,a_t)=\mathbb{E}
\left[r_t+\gamma \underset{a'}{\mathrm{max}}Q^*(s_{t+1},a')
\right]
\end{equation*}

\subsection{Deep Q Networks} \citep{mnih2015human,mnih2013playing} approximate the optimal Q function using a Convolutional Neural Network (CNN). The training objective is to minimize the expected TD error of the optimal Bellman equation:
\begin{equation*}
\mathbb{E}_{s_t,a_t,r_t,s_{t+1}}\left\Vert Q_{\theta}\left(s_{t},a_{t}\right)-y_{t}\right\Vert _{2}^{2}
\end{equation*}
\begin{equation*}
y_{t}=
\begin{cases}
r_t & s_{t+1} \mbox{ is terminal}\\
r_{t}+\gamma\underset{\mbox{\mbox{\ensuremath{a}'}}}{\mbox{max}}Q_{\theta_{target}}\left(s_{t+1},a^{'}\right) & \mbox{otherwise}
\end{cases}
\end{equation*}

Notice that this is an off-line algorithm, meaning that the tuples $\left\{ s_{t,}a_{t},r_{t},s_{t+1},\gamma\right\}$ are collected from the agents experience, stored in the ER and later used for training. The reward $r_t$ is clipped to the range of $[-1,1]$ to guarantee stability when training DQNs over multiple domains with different reward scales. The DQN algorithm maintains two separate Q-networks: one with parameters $\theta$, and a second with parameters $\theta_{target}$ that are updated from $\theta$ every fixed number of iterations. In order to capture the game dynamics, the DQN algorithm represents a state by a sequence of history frames and pads initial states with zero frames.\\

\subsection{Visualization}

The goal behind visualization of DNNs is to give a better understanding of these inscrutable black boxes. While some approaches study the type of computation performed at each layer as a group \citep{ yosinski2014transferable}, others try to explain the function computed by each individual neuron (similar to the "Jennifer Aniston Neuron" \citep{quiroga2005invariant}). Dataset-centric approaches display images from the data set that cause high activations of individual units. For example, the deconvolution method \citep{zeiler2014visualizing} highlights the areas of a particular image that are responsible for the firing of each neural unit. Network-centric approaches investigate a network directly without any data from a dataset, e.g., \citep{ erhan2009visualizing} synthesized images that cause high activations for particular units. Other works used the input gradient to find images that cause strong activations (e.g., \citep{simonyan2014very}; \citep{nguyen2014deep}; \citep{szegedy2013intriguing}).\\

Since RL research had been mostly focused on linear function approximations and policy gradient methods, we are less familiar with visualization techniques and attempts to understand the structure learned by an agent. \citep{wang2015dueling}, suggested to use saliency maps and analyzed which pixels affect network predictions the most. Using this method they compared between the standard DQN and their dueling network architecture. \citep{engel2001learning} learned an embedded map of Markov processes and visualized it on two dimensions. Their analysis is based on the state transition distribution while we will focus on distances between the features learned by DQN.\\

\textbf{t-SNE} is a non-linear dimensionality reduction method used mostly for visualizing high dimensional data. The technique is easy to optimize, and it has been proven to outperform linear dimensionality reduction methods and non-linear embedding methods such as ISOMAP \citep{tenenbaum2000global} in several research fields including machine learning benchmarks and hyper-spectral remote sensing data \citep{lunga2014manifold}. t-SNE reduces the tendency to crowd points together in the center of the map by employing a heavy-tailed Student-t distribution in the low dimensional space. It is known to be particularly good at creating a single map that reveals structure at many different scales, which is particularly important for high-dimensional data that lies on several low-dimensional manifolds. This enables us to visualize the different sub-manifolds learned by the network and interpret their meaning.\\

%
%The \textbf{Skill Policy} $\mu : S\rightarrow \Delta_\Sigma$  is a mapping from states to a probability distribution over skills. The action-value function $Q_\mu(s, \sigma) = \mathbb{E} [\sum ^\infty _{t=0} \gamma ^t R_t |(s, \sigma), \mu] $ represents the value of choosing skill $\sigma \in \Sigma$ at state $s \in S$, and thereafter selecting skills according to policy $\mu$.
%The optimal skill value function is given by: $
%\label{OptionBellman}
%Q_{\Sigma}^*(s,\sigma) = \mathbb{E} [R_s^{\sigma} + \gamma ^k \underset{\sigma'\in \Sigma}{\mathrm{max}} Q_{\Sigma}^*(s',\sigma')] \enspace$ \citep{stolle2002learning}.

\subsection{Abstractions}
An abstraction is an important tool that enables an agent to focus less on the lower level details of a task and more on solving the task at hand. Many real-world domains can be modeled using some form of abstraction. The Semi MDP (SMDP) and the Aggregated MDP (AMDP) models are two important examples that extend the standard MDP formulation by using temporal and spatial abstractions respectively.\\ 

\textbf{SMDP.} \citep{sutton1999between}, extends the MDP action space $A$ and allows the agent to plan with skills $\Sigma$ (also known as options, macro-actions and temporally-extended actions) . A skill is defined by a triplet: $\sigma = <I,\pi,\beta>,$ where $I$ defines the set of states where the skill can be initiated, $\pi$ is the intra-skill policy, and $\beta$ is the set of termination probabilities determining when a skill will stop executing. $\beta$ is typically either a function of state $s$ or time $t$. Planning with skills can be performed by learning for each state the value of choosing each skill. More formally, an SMDP can be defined by a five-tuple $<S, \Sigma, P, R, \gamma>,$ where $S$ is the set of states, $\Sigma$ is the set of skills, $P$ is the SMDP transition matrix, $\gamma$ is the discount factor and the SMDP reward is defined by:
\begin{equation}
\label{eq:skill_reward}
R_s^{\sigma} = \mathbb{E}[r_s^{\sigma}] = \mathbb{E}[r_{t+1} + \gamma r_{t+2} + \cdot\cdot\cdot + \gamma ^{k-1} r_{t+k} | s_t=s,\sigma] 
\end{equation}

Analyzing policies using the SMDP model shortens the planning horizon and simplifies the analysis. However, there are two drawbacks with this approach. First, one must consider the high complexity of the state space, and second, the SMDP model requires to identify a set of skills, a challenging problem with no easy solution \citep{mankowitz2016adaptive}.\\

\textbf{AMDP.} Another approach is to analyze a policy using spatial abstractions in the state space. If there is a reason to believe that groups of states share common attributes such as similar policy or value function, it is possible to use State Aggregation \citep{moore-variableresolution}. State Aggregation is a well-studied problem that typically involves identifying clusters as the new states of an AMDP, where the set of clusters $C$ replaces the MDP states $S$. Doing RL on aggregated states is potentially advantageous because the transition probability matrix $P$, the reward signal $R$ and the policy $\pi$ dimensions are decreased, e.g., \citep{singh1}. The AMDP modeling approach has two drawbacks. First, the action space $A$ is not modified and therefore the planning horizon is still intractable. Second, AMDPs are not necessarily Markovian \citep{baimarkovian}.\\

\section{Methods}
\subsection{Manual clustering}
\begin{figure}
\vspace{-0.5cm}
\centerline{\includegraphics[trim=0cm 0cm 0cm 0.5cm,clip, width=0.8\textwidth]{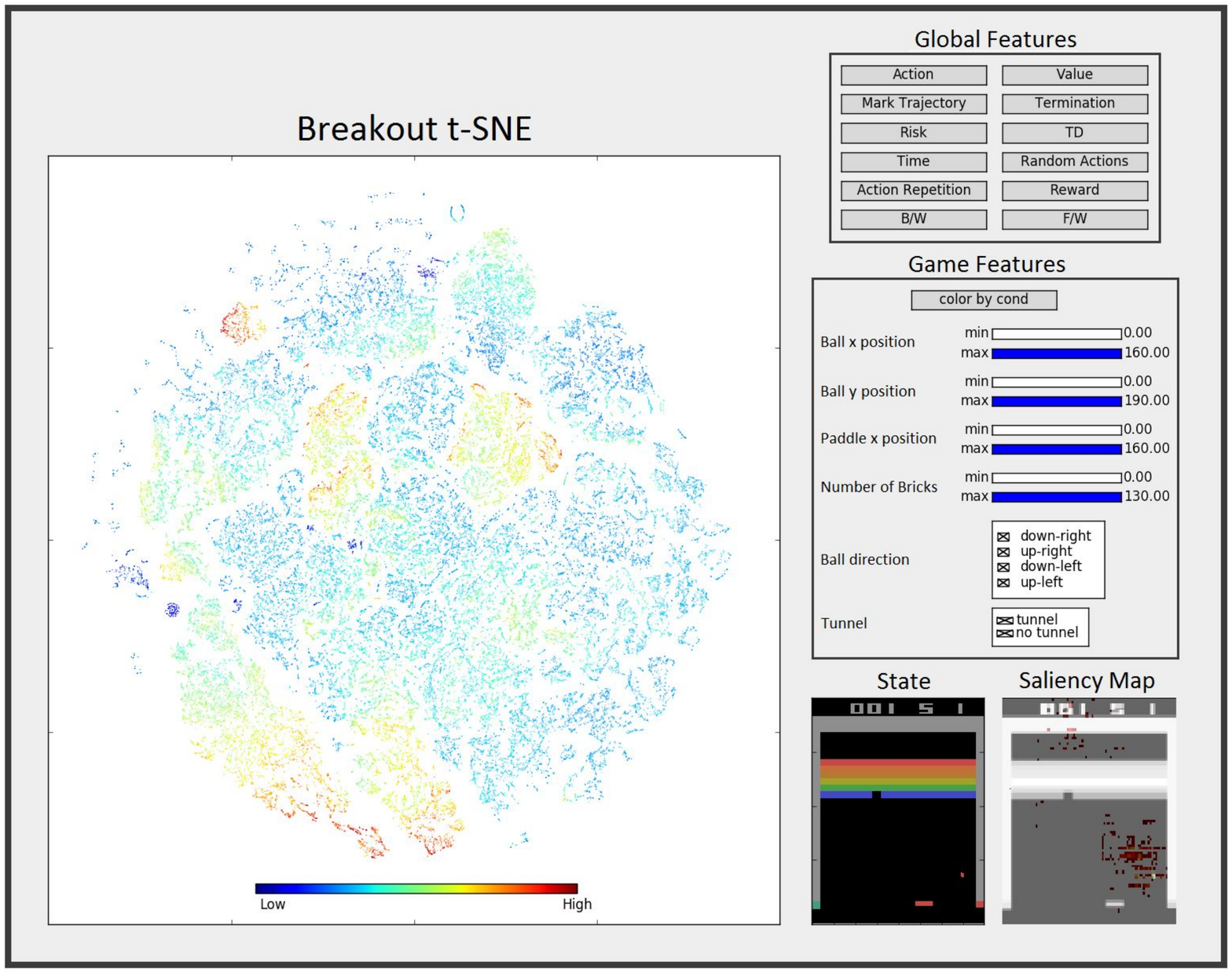}}
\caption{Graphical user interface for our methodology.}
\label{Tool_description}
\vspace{-0.5cm}
\end{figure}

Figure~\ref{Tool_description} presents our Visualization tool. Each state is represented as a point in the t-SNE map (Left). The color of the points is set manually using global features (Top right) or game specific hand crafted features (Middle right). Clicking on each data point displays the corresponding game image and saliency map (Bottom right). It is possible to move between states along the trajectory using the F/W and B/W buttons.\\

\textbf{Feature extraction:} During the DQN run, we collect statistics for each state such as Q values estimates, generation time, the termination signal, reward and played action. We also extract hand-crafted features, directly from the emulator raw frames, such as player position, enemy position, the number of lives, and so on. We use these features to color and filter points on the t-SNE map. The filtered images reveal insightful patterns that cannot be seen otherwise. From our experience, hand-crafted features are very powerful, however, the drawback of using them is that they require manual work. In all the figures below we use a heat color map (red corresponds to high values and blue to low ones).\\

Similar to \citep{engel2001learning} we visualize the dynamics (state transitions) of the learned policy. To do so we use a 3D t-SNE state representation wich we found insightfull. The transitions are displayed with arrows using Mayavi \citep{ramachandran2011mayavi}.\\

\textbf{t-SNE:}
We apply the t-SNE algorithm directly on the collected neural activations, similar to \citep{mnih2015human}. The input $X\in\mathbf{R}^{120k\times512}$ consists of $120k$ game states with $512$ features each (size of the last layer). Since these data are relatively large, we pre-processed it using Principal Component Analysis to dimensionality of 50 and used the Barnes-Hut t-SNE approximation \citep{van2014accelerating}.\\

\textbf{Saliency maps:}
We generate Saliency maps (similar to \citep{simonyan2014very}) by computing the Jacobian of the network with respect to the input and presenting it above the input image itself (Figure~\ref{Tool_description}, bottom right). These maps helps to understand which pixels in the image affect the value prediction of the network the most.\\ 

\textbf{Analysis:}
Using these features we are able to understand the common attributes of a given cluster (e.g, Figure ~\ref{SeaquestFiltering} in the appendix). Moreover, by analyzing the dynamics between clusters we can identify a hierarchical aggregation of the state space. We define clusters with a clear entrance and termination areas as options and interpret the agent policy there. For some options we are able to derive rules for initiation and termination (i.e., landmark options, \citep{sutton1999between,mann2015approximate}).\\\\
To summarize, the manual clustering methodology comprises the following steps:
\begin{enumerate}
\item Train a DQN.
\item Evaluate the DQN. For each visited state record: (a) activations of the last hidden layer, (b) the gradient information and (c) the game state (raw pixel frame).
\item Apply t-SNE on the activations data.
\item Visualize the data. 
\item Analyze the policy manually.
\end{enumerate}

\begin{figure}[t]
\begin{tabular}{ c @{\hskip 0.35in} c }
   \includegraphics[width=0.25\textwidth, height=0.32\textwidth]{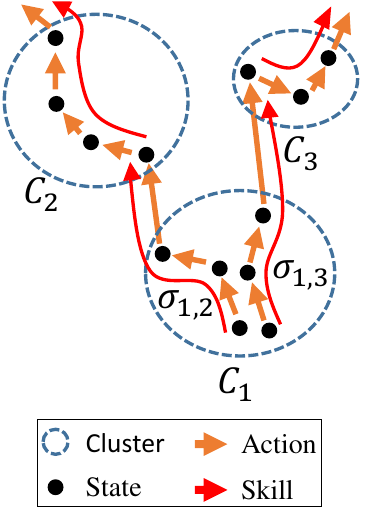} 

 &  \includegraphics[width=0.65\textwidth, height=0.32\textwidth]{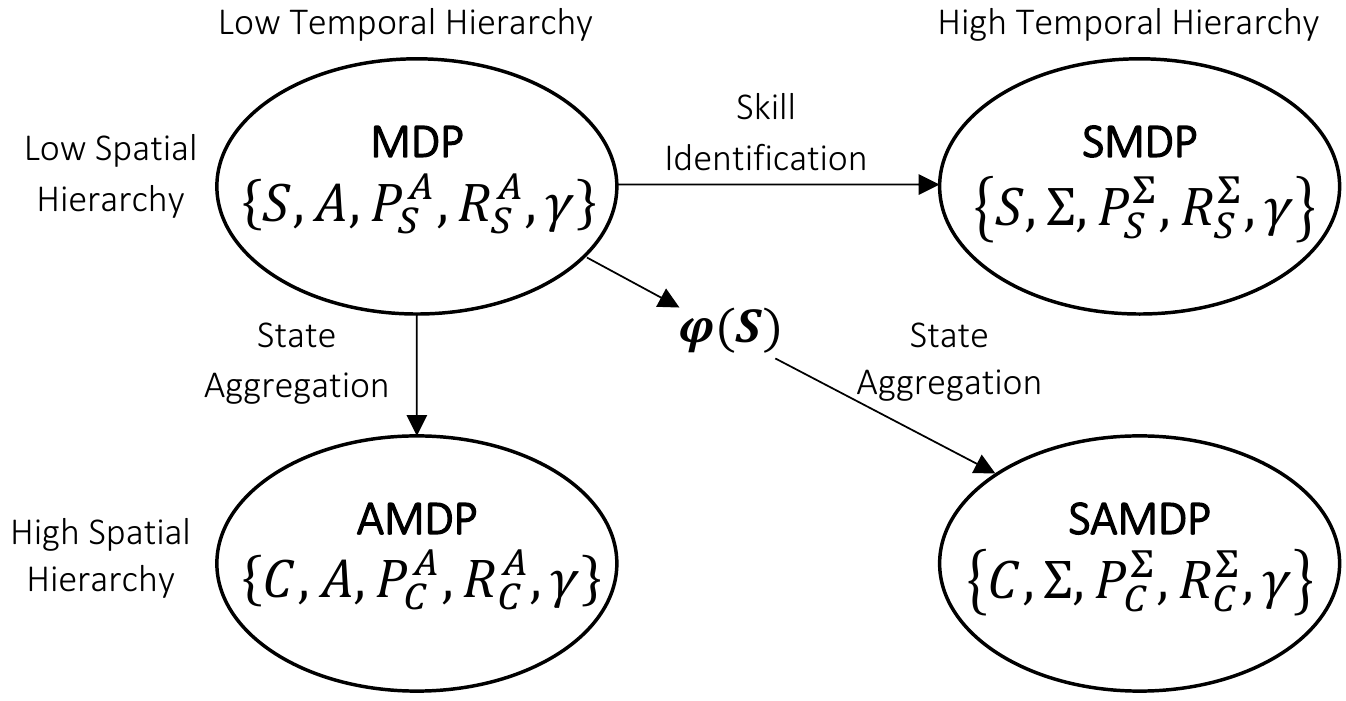} 
\end{tabular}
\caption{\textbf{Left:} Illustration of state aggregation and skills. Primitive actions (orange arrows) cause transitions between MDP states (black dots) while skills (red arrows) induce transitions between SAMDP states (blue circles). \textbf{Right:} Modeling approaches for analyzing policies.}
\label{fig:modeling}
\end{figure}

\subsection{The Semi Aggregated MDP}
\label{sec:samdp}

In this paper, we propose a model that combines the advantages of the SMDP and AMDP approaches and denote it by \textbf{SAMDP}. Under SAMDP modeling, aggregation defines both the states and the set of skills, allowing analysis with spatio-temporal abstractions (the state-space dimensions and the planning horizon are reduced). However, SAMDPs are still not necessarily Markovian. We summarize the different modeling approaches in Figure \ref{fig:modeling}. The rest of this section is devoted to explaining the five stages of building an SAMDP model: (0) Feature selection, (1) State Aggregation, (2) Skill identification, (3) Inference, and (4) Model Selection.\\

\textbf{(0) Feature selection.} We define the mapping from MDP states to features, by a mapping function $\phi:s\rightarrow s' \subset R^m $. The features may be raw (e.g., spatial coordinates, frame pixels) or higher level abstractions (e.g., the last hidden layer of a NN). The feature representation has a significant effect on the quality of the resulting SAMDP model and visa versa, good model can point out a good feature representation.\\

\textbf{(1) Aggregation via Spatio-temporal clustering.} The goal of Aggregation is to find a mapping (clustering) from the MDP feature space $S' \subset R^m$ to the AMDP state space $C$. Typically, clustering algorithms assume that the data is drawn from an i.i.d distribution. However, in our problem the data is generated from an MDP and therefore violates this assumption. We alleviate this problem using two different approaches. First, we decouple the clustering step with the SAMDP model. This is done by creating an ensemble of clustering candidates and building an SAMDP model for each (following stages 2 and 3). In stage 4, we will explain how to run a non-analytic outer optimization loop to choose between these candidates based on spatio-temporal evaluation criteria. Second, we introduce a novel extension of the celebrated K-means algorithm \citep{macqueen1967some}, that enforces temporal coherency along trajectories. In the vanilla K-means algorithm, a cluster $c_i$ with mean $\mu_i$ is assigned to a point $x_t$ such that the cluster with the closest mean to the point $x_t$ is chosen (please refer to the supplementary material for more details). We modified this step as follows: $$ c(x_t) = \big \{ c_i : \big \| X_t - \mu_i \big \|^2_F \le \big \| X_t - \mu_j \big \|^2_F, \forall j \in [1,K] \big\}.$$ Here, F stands for the Frobenius norm, K is the number of clusters, $t$ is the time index of $x_t$, and $X_t$ is a set of $2w+1$ centered at $x_t$ from the same trajectory: $\big \{ x_j \in X_t \iff j \in [t-w,t+w] \big\}$. The dimensions of $\mu$ correspond to a single point, but it is expanded to the dimensions of $X_t$. In this way, a point $x_t$ is assigned to a cluster $c_i$, if its neighbors along the trajectory are also close to $\mu_i$. Thus, enforcing temporal coherency.\\
We have also experienced with other clustering methods such as spectral clustering, hierarchical agglomerative clustering and entropy minimization (please refer to the supplementary material for more details).\\

\textbf{(2) Skill identification.} We define an SAMDP \textbf{skill} $\sigma_{i,j}\in \Sigma$ uniquely by a single initiation state $c_i \in C$ and a single termination state $c_j \in C: \sigma_{ij} = < c_i ,\pi_{i,j}, c_j >.$
More formally, at time $t$ the agent enters an AMDP state $c_i$ at an MDP state $s_t \in c_i$. It chooses a skill according to its SAMDP policy and follows the skill policy $\pi_{i,j}$ for $k$ time steps until it reaches a state $s_{t+k} \in c_j$, s.t $i \neq j$. We do not define the skill length $k$ a-priori nor the skill policy but infer the skill length from the data. As for the skill policies, our model does not define them explicitly but we will observe later that our model successfully identifies skills that are localized in time and space.\\

\textbf{(3) Inference.} Given the SAMDP states and skills, we infer the skill length, the SAMDP reward and the SAMDP probability transition matrix from observations. The \textbf{skill length}, is inferred for a skill $\sigma_{i,j}$ by averaging the number of MDP states visited since entering SAMDP state $c_i$ and until leaving for SAMDP state $c_j$. The \textbf{skill reward} is inferred similarly using Equation~\ref{eq:skill_reward}.\\
The inference of the SAMDP \textbf{transition matrices} is a bit more puzzling, since the probability of seeing the next SAMDP state depends both on the MDP dynamics and the agent policy in the MDP state space. We now turn to discuss how to infer these matrices by observing transitions in the MDP state space. 
Our goal is to infer two quantities: (a) The SAMDP \textbf{transition probability matrices} $P_\Sigma: P^{\sigma \in \Sigma}_{i,j}=Pr(c_j|c_i,\sigma)$, measures the probability of moving from state $c_i$ to $c_j$ given that skill $\sigma$ is chosen. These matrices are defined uniquely by our definition of skills as deterministic probability matrices.  (b) The probability of moving from state $c_i$ to $c_j$ given that skill $\sigma$ is chosen according to the agent SAMDP policy: $P^{\pi}_{i,j}=Pr(c_j|c_i,\sigma=\pi(c_i))$. This quantity involves both the SAMDP \textbf{transition probability matrices} and the agent policy. However, since SAMDP transition probability matrices are deterministic, this is equivalent to the agent policy in the SAMDP state space. Therefore by inferring transitions between SAMDP states we directly infer the agent's SAMDP policy. \\
Given an MDP with a deterministic environment and an agent with a nearly deterministic MDP policy (e.g., a deterministic policy that uses an $\epsilon$-greedy exploration ($\epsilon \ll 1$)), it is intuitive to assume that we would observe a nearly deterministic SAMDP policy. However, there are two sources that increase the stochasticity of the SAMDP policy:
(1) Stochasticity is accumulated along skill trajectories. (2) The aggregation process is only an approximation. A given SAMDP state may contain more than one "real" state and therefore more than one skill. Performing inference, in this case, we will simply observe a stochastic policy that chooses between those skills at random.\\
Thus, it is very likely to infer a stochastic SAMDP transition matrix, although the SAMDP transition probability matrices are deterministic and even if the MDP environment is deterministic and the MDP policy is near deterministic.\\

\textbf{(4) Model selection.}\label{sub:eval} So far we explained how to build an SAMDP from observations. In this stage we explain how to choose between different SAMDP model candidates. There are two advantages for choosing between multiple SAMDPs. First, there are different hyper parameter to tune. Two examples are the number of SAMDP states (K) and the window size (w) for the clustering algorithm. Second, there is randomness in the aggregation step. Hence, clustering multiple times and picking the best result will potentially yield better models.\\
We therefore developed evaluation criteria that allow us to select the best model, motivated by \cite{hallak2013model}. We follow the Occam’s Razor principle, and aim to find the simplest model which best explains the data. \textit{(i)} \textbf{Value Mean Square Error(VMSE)}, measures the consistency of the model with the observations. The estimator is given by
%\begin{equation}
%\label{eq:vmse}
%\mbox{VMSE} = \frac{\| v-v_{SAMDP} \|}{\|v\|},
%\end{equation}
\begin{equation}
\label{eq:vmse}
VMSE = \frac{\| v-v_{SAMDP} \|}{\|v\|}
\end{equation}
where $v$ stands for the SAMDP value function of the given policy, and $v_{SAMDP}$ is given by: $ v_{SAMDP} = ( I+\gamma^{k}P )^{-1}r $, where P is measured under the SAMDP policy.
\textit{(ii)} \textbf{Inertia}, the K-means algorithm objective function, is given by : $I = \sum_{i=0}^{n}\min_{\mu_j \in C}(||x_j - \mu_i||^2).$ Inertia measures the variance inside clusters and encourages spatial coherency. Motivated by Ncut and spectral clustering \citep{von2007tutorial}, we define \textit{(iii)} \textbf{The Intensity Factor} as the fraction of out/in cluster transitions. However, we define edges between states that are connected along the trajectory (a transition between them was observed) and give them equal weights (instead of defining the edges by euclidean distances as in spectral clustering). Minimizing the intensity factor encourages longer duration skills. \textit{(iv)} \textbf{Entropy}, is defined on the SAMDP probability transition matrix as follows: $e= - \sum_i \{ |C_i| \cdot \sum_j{P_{i,j} \log P_{i,j}} \}.$ Low entropy encourages clusters to have less skills, i.e., clusters that are localized both in time and space.

\section{Experiments}

In this section, we demonstrate our method on a Gridword domain as well as on three ATARI games: Pacman (a game where DQN performs very well), Seaquest (from the opposite reason) and Breakout (a popular game). For each domain we: (1) provide a short description of the domain and the optimal policy, (2) detail the hand-crafted features we designed for the domain, (3) analyze the DQN's policy using the manual approach and the SAMDP model and derive conclusions. We then analyze two cross-domain observations: the representation of initial and terminal states and the influence of score pixels in Atari. We finish with additional experiments on the SAMDP model. We show that it is consistent and suggest a novel shared autonomy application.

\subsection{Gridworld}
\begin{figure}
\vspace{-1cm}
  \centering
  \subfigure[MDP]{\includegraphics[scale=0.26]{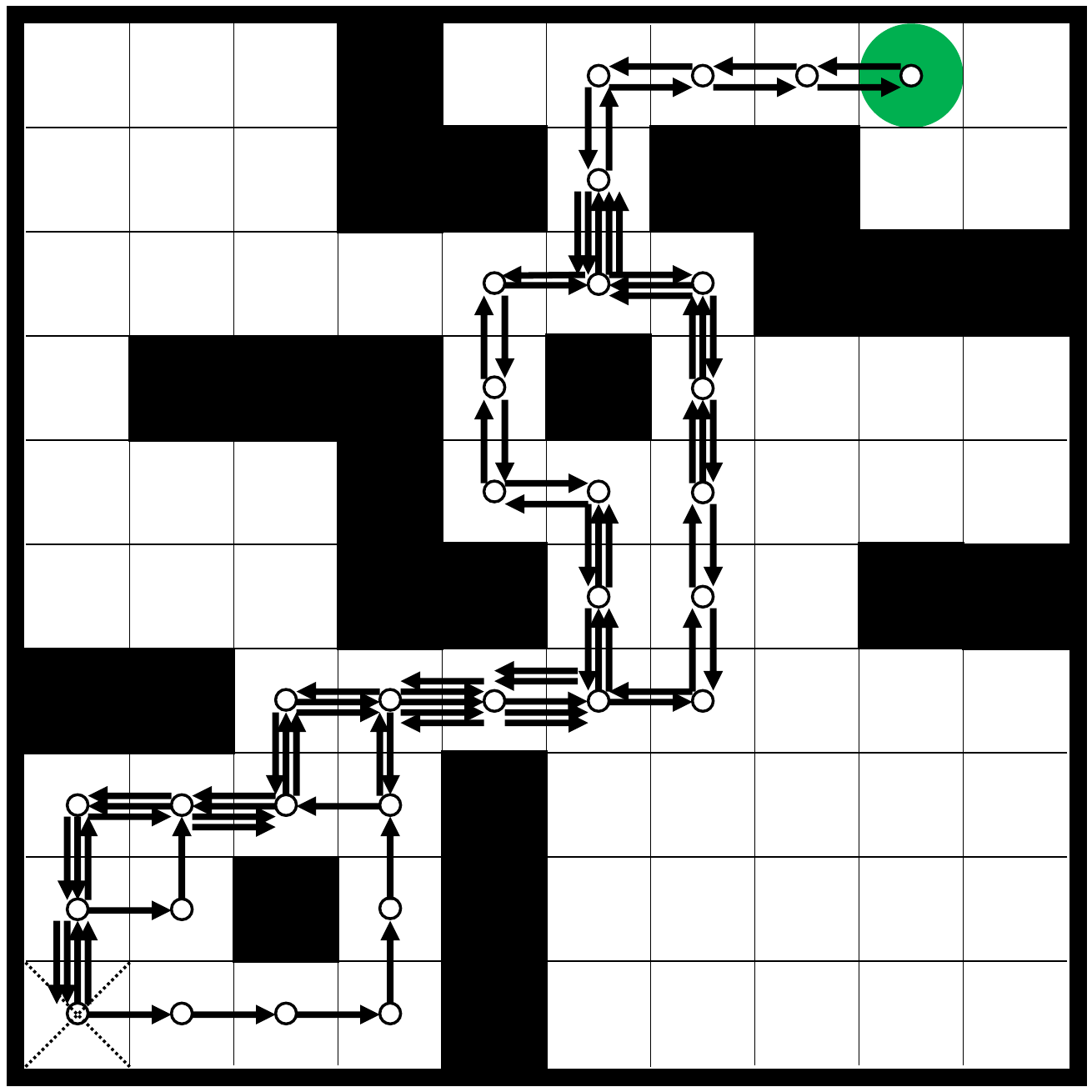}\label{gridworld_mdp}}
  \subfigure[SMDP]{\includegraphics[scale=0.26]{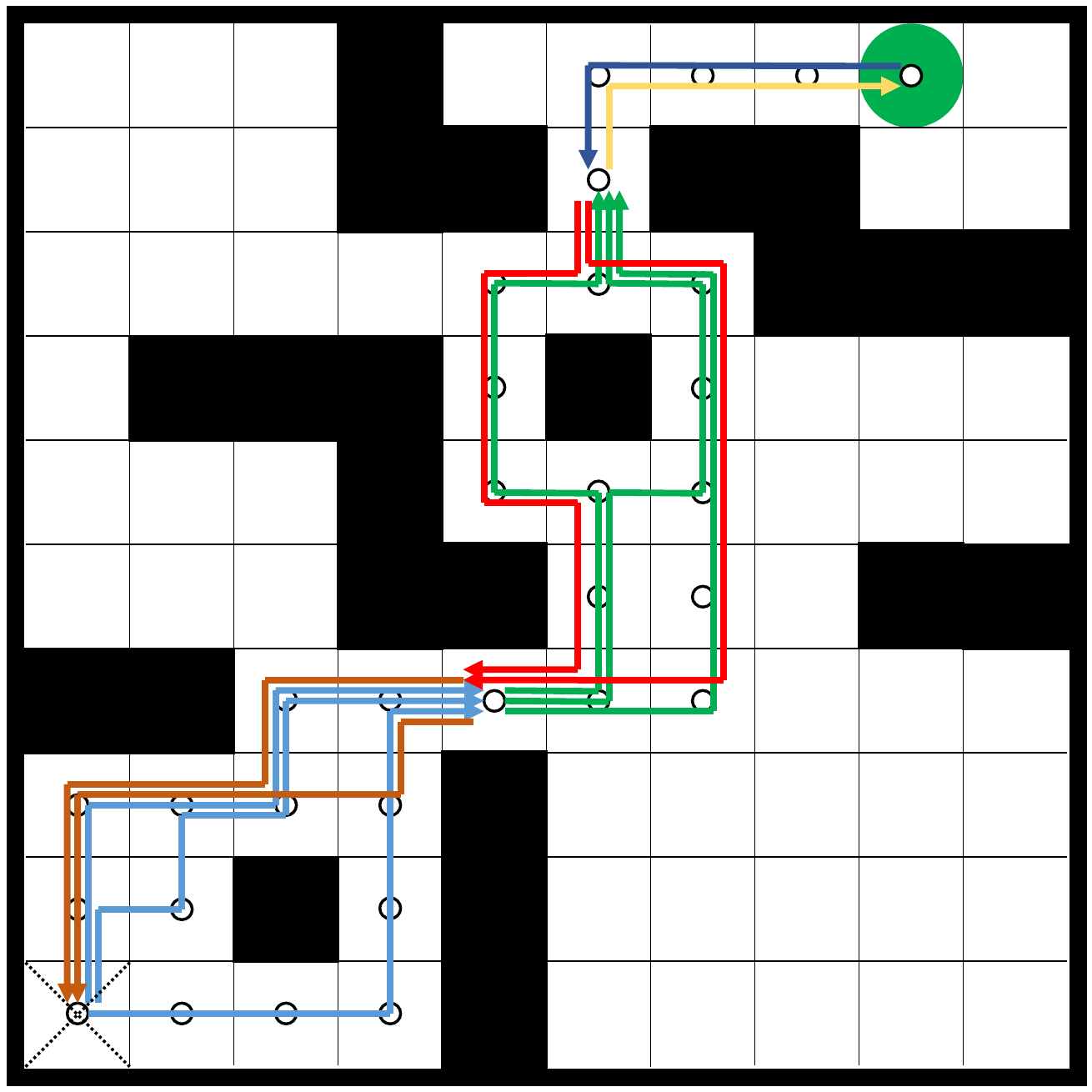}\label{gridworld_smdp}}
  \subfigure[AMDP]{\includegraphics[scale=0.26]{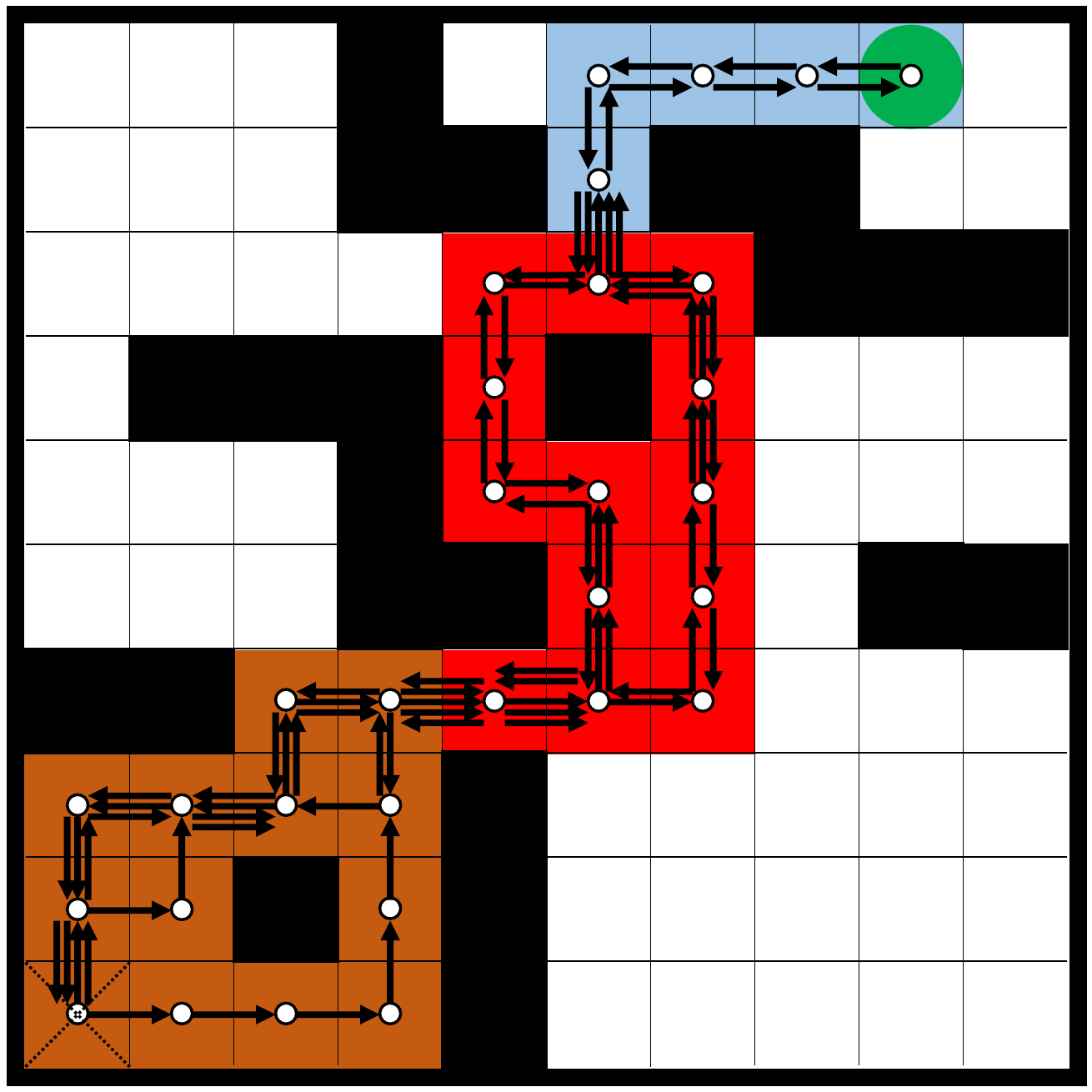}\label{gridworld_amdp}}
  \subfigure[SAMDP]{\includegraphics[scale=0.26]{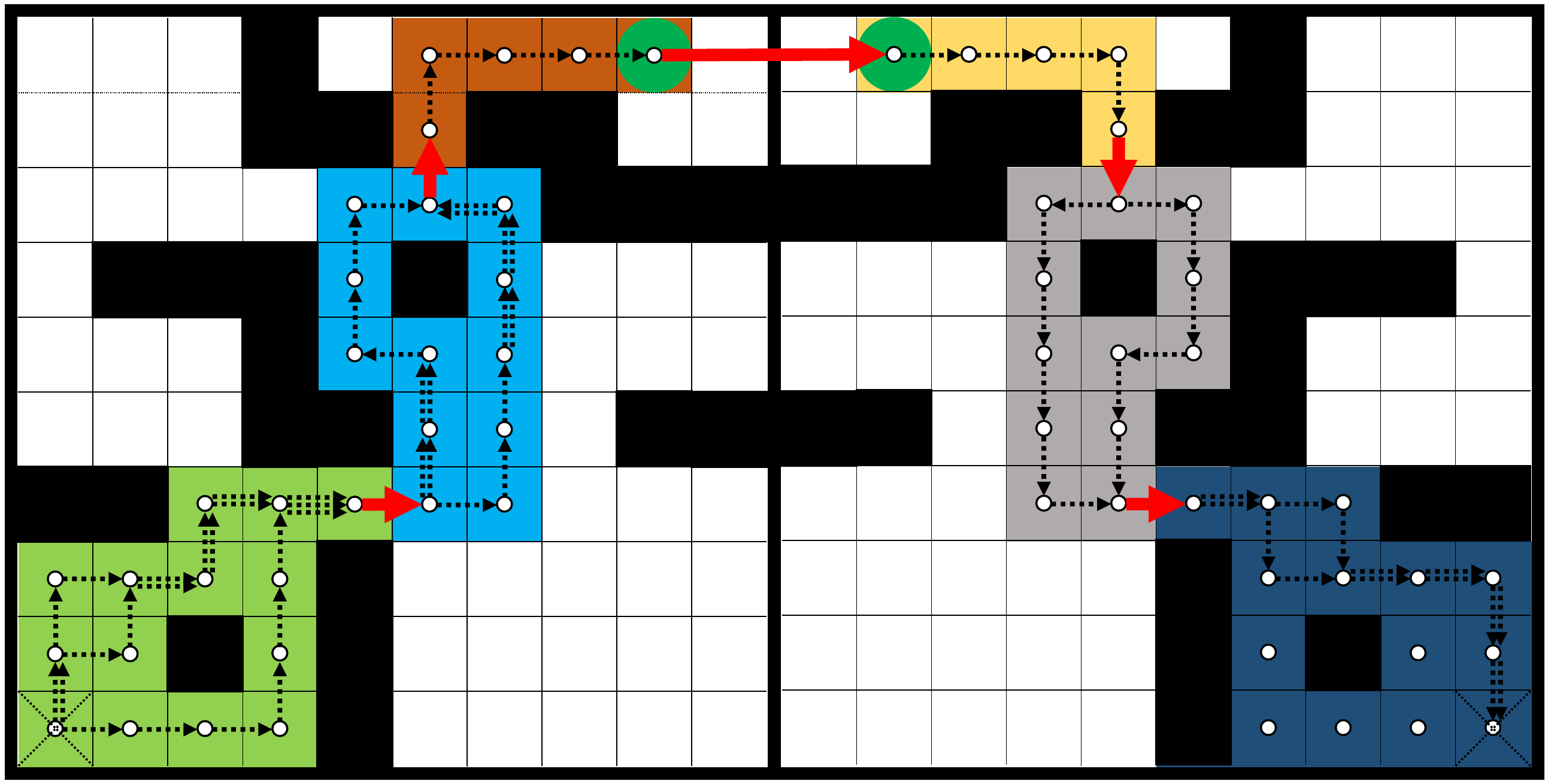}\label{gridworld_samdp}}
  \caption{State-action diagrams for a gridworld problem. \textbf{a.} \textbf{MDP} diagram: relate to individual states and primitive actions. \textbf{b.} \textbf{SMDP} diagram: Edge colors represent different skills. \textbf{c.} \textbf{AMDP} diagram: clusters are formed using spatial aggregation in the original state. \textbf{d.} \textbf{SAMDP} diagram: clusters are found after transforming the state space. intra-cluster transitions (dashed arrows) can be used to explain the skills, while inter-cluster transitions (big red arrows) loyaly explain the governing policy.}
  \label{toy_problem}
  \vspace{-0.6cm}
\end{figure}
\textbf{Description.} An agent is placed at the origin (marked in X), where the goal is to reach the green ball and return. \textbf{Hand-crafted features.} The state $s\in\emph{R}^3$ is given by: $s=\{x,y,b\}$, where $(x,y)$ are the coordinates and $b\in \{0,1\}$ indicates whether the agent has reached the ball or not. The policy is trained to find skills following \cite{Mann2014b}. \textbf{Analysis.} We visualize the learned policy using the maze coordinates with the different modeling approaches. The MDP graph (Figure~\ref{toy_problem}a), consists of a large number of states. It is also difficult to understand which skills the agent is using. In the SMDP graph (Figure~\ref{toy_problem}b), the number of states remain high, however coloring the edges by the skills, helps to understand the agent's behaviour. Unfortunately, producing this graph is seldom possible because we rarely receive information about the skills. On the other hand, abstracting the state space can be done more easily using state aggregation. However, in the AMDP graph (Figure~\ref{toy_problem}c), the clusters are not aligned with the played skills because the routes leading \textit{to} and \textit{from} the ball overlap.\\
\textbf{SAMDP.} We use the following state representation:
$$\phi(x,y) = \begin{cases} (x,y), & \mbox{if } b\mbox{ is 0} \\ (2L-x,y), & \mbox{if } b\mbox{ is 1} \end{cases}$$
where $L$ is the maze width. The transformation $\phi$ flips and translates the states where $b=1$. Now that the routes \textit{to} and \textit{from} the ball are disentangled, the clusters are aligned with the skills. Understanding the behaviour of the agent is now possible by examining inter-cluster and intra-cluster transitions (Figure~\ref{toy_problem}d). The key observation from the gridworld domain is that under the correct representation, the hirerchy of the MDP is discovered by the SAMDP model. In the next experiments we will show that DQN is capable to learn such a representation automatically.

\subsection{Setup for Atari}
\textbf{Data collection.} We record the neural activations of the last hidden layer and the DQN value estimations for 120k game states (indexed by visitation order). \textbf{t-SNE} is applied on the neural activations data, a non-linear dimensionality reduction method that is particularly good at creating a single map that reveals structure at many different scales. The result is a compact, well-separated representation, that is easy and fast to cluster. \\\\
\textbf{SAMDP} algorithm details. \\
\textbf{Feature selection.} We use the t-SNE coordinates and a value coordinate (three coordinates in total). Each coordinate is normalized to have zero mean and unit variance. We also experimented with other configurations such as using the activations without t-SNE and different normalization. However, we found that this configuration is fast and gives nice results. \textbf{SAMDP clustering and skill identification} follow Section~\ref{sec:samdp} with no additional details. \textbf{SAMDP inference.} We use two approximations in the inference stage which we found to work well: we overlook transitions with small skill length (shorter than 2) and we truncate transitions that have less than 0.1 probability.
\textbf{SAMDP model Selection.} We perform a grid search on two parameters: \textit{i}) number of clusters \textbf{$K \in [15,25]$}. \textit{ii}) window size \textbf{$w \in [1,7]$}. We found that models larger (smaller) than that are too cumbersome (simple) to analyze. We select the best model in the following way: we first sort all models by the four evaluation criteria (SAMDP Section, stage 4) from best to worst. Then, we iteratively intersect the p-prefix of all sets (i.e., the first p elements of each set) starting with 1-prefix. We stop when the intersection is non-empty and choose the configuration at the intersection. We also measure the p-value of the chosen model. For the null hypothesis, we choose the SAMDP model constructed with random clusters. We tested 10000 random SAMDP models, none of which scored better than the chosen model (for all the evaluation criteria).\\

\newpage
\subsection{Breakout}

\textbf{Description.} In \href{https://archive.org/details/ArcadeGameManualBreakout}{Breakout}, a layer of bricks lines the top of the screen. A ball travels across the screen, bouncing off the top and side walls. When a brick is hit, the ball bounces away, the brick is destroyed and the player receives a reward. The player loses a turn when the ball touches the bottom of the screen. To prevent this from happening, the player has a movable paddle to bounce the ball upward, keeping it in play. The highest score achievable for one player is 896; this is done by eliminating exactly two screens of bricks worth 448 points each. A good strategy for Breakout is leading the ball above the bricks by digging a tunnel on one side of the bricks block. Doing so enables the agent to achieve high reward while being safe from losing the ball. By introducing a discount factor to Breakout's MDP, this strategy becomes even more favorable since it achieves high immediate reward.\\

\begin{figure}
\begin{center}
\centerline{\includegraphics[trim=2cm 4cm .5cm 0cm,clip,width=0.8\textwidth]{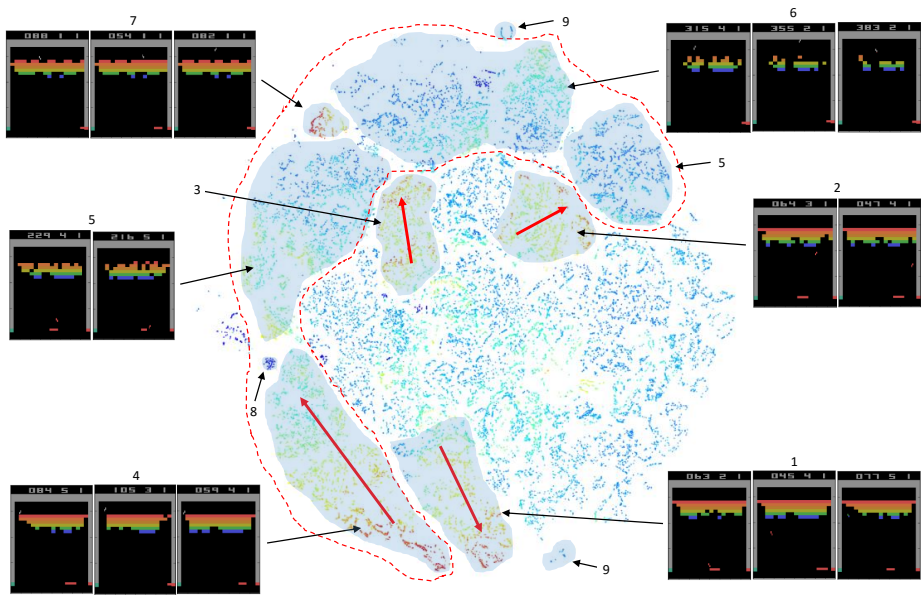}} 
\caption{Breakout aggregated states on the t-SNE map.}
\label{Breakout_Aggregated}
\end{center}
\end{figure} 

\textbf{Hand-crafted features.} We extract features for the player (paddle) position, ball's position, ball’s direction of movement, number of missing bricks, and a tunnel feature (a tunnel exists when there is a clear path between the area below the bricks and the area above it, and we approximate this event by looking for at least one clear column of bricks).\\

\textbf{Analysis.} Figure~\ref{Breakout_Aggregated} presents the t-SNE of Breakout. The agent learns a hierarchical policy: (a) carve a tunnel in the left side of the screen and (b) keep the ball above the bricks as long as possible. In clusters (1-3) the agent is carving the left tunnel. Once the agent enters those clusters, it will not exit until the tunnel is carved (see Figure~\ref{BreakoutOption}). We identify these clusters as a landmark option. The clusters are separated from each other by the ball position and direction. In cluster 1 the ball is heading toward the tunnel, in cluster 2 the ball is on the right side and in cluster 3 the ball bounces back from the tunnel after hitting bricks.  As fewer bricks remain in the tunnel the value is gradually rising till the tunnel is carved where the value is maximal (this makes sense since the agent is enjoying high reward rate straight after reaching it). Once the tunnel is carved, the option is terminated and the agent moves to clusters 4-7 (dashed red line), differentiated by the ball position with regard to the bricks (see Figure~\ref{Breakout_Aggregated}). In cluster 4 and 6 the ball is above the bricks and in 5 it is below them. Clusters 8 and 9 represent termination and initial states respectively (see Figure ~\ref{BreakoutOption} in the appendix for examples of states along the option path). \\

\begin{figure}
\centering
    \begin{subfigure}

        \includegraphics[trim=3cm 18cm 6cm 3cm,clip,width=8cm]{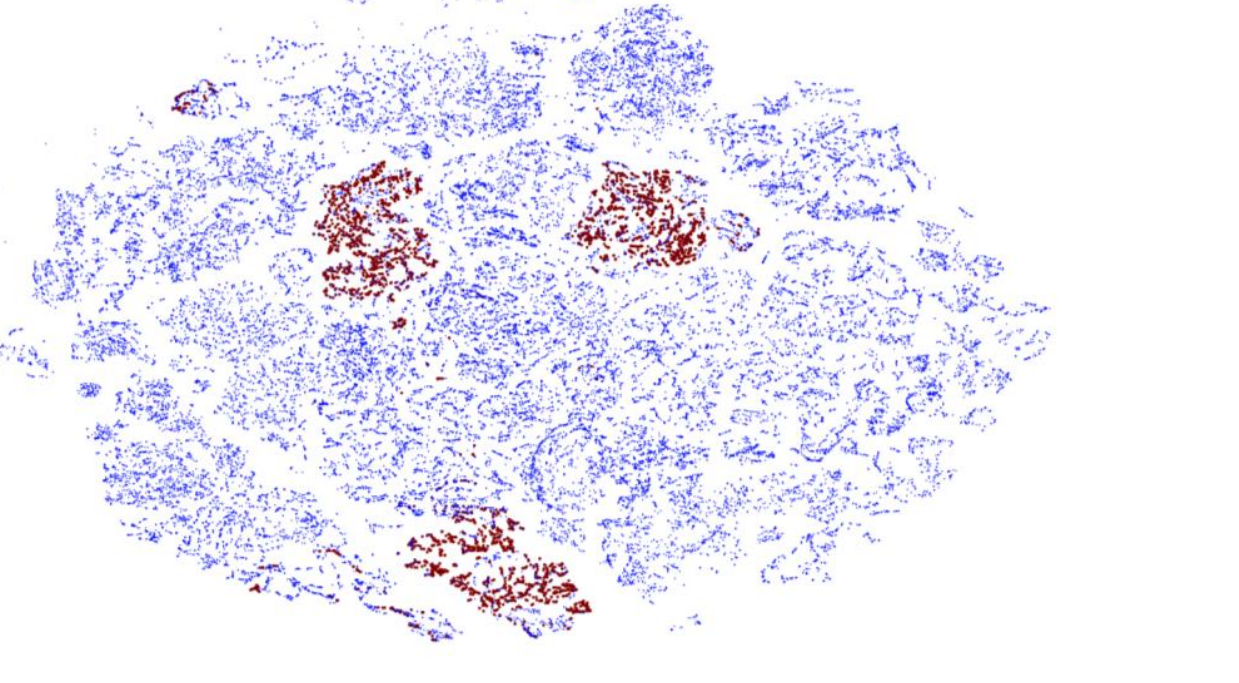}
    \end{subfigure}
    \begin{subfigure}
        
        \includegraphics[trim=1cm 0cm 1cm 0cm,clip,width=5cm]{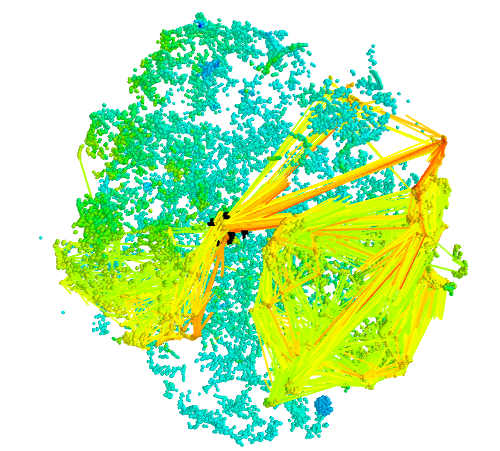}
    \end{subfigure}
\caption{Breakout tunnel digging option. \textbf{Left:} states that the agent visits once it entered the option clusters (1-3 in Figure~\ref{Breakout_Aggregated}) until it finishes to carve the left tunnel are marked in red. \textbf{Right:} Dynamics is displayed by arrows above a 3d t-SNE map. The option termination zone is marked by a black annotation box and corresponds to carving the left tunnel. All transitions from clusters 1-3 into clusters 4-7 pass through a singular point.}
\label{BreakoutOption}
\vspace{-0.5cm}
\end{figure}

Cluster 7 is created due to a bug in the emulator that allows the ball to pass the tunnel without completely carving it. The agent learned to represent this incident to its own cluster and assigned it high-value estimates (same as the other tunnel clusters). This observation is interesting since it indicates that the agent is learning a representation based on the game dynamics and not only on the pixels.\\

By coloring the t-SNE map by time, we can identify some policy downsides. States with only a few remaining bricks are visited for multiple times along the trajectory (see Figure ~\ref{BreakoutTime} in the appendix). In these states, the ball bounces without hitting any bricks which causes a fixed reflection pattern, indicating that the agent is stuck in local optima. We discuss the inability of the agent to perform well on the second screen of the game in Section~\ref{score_section}.\\

\begin{figure}
\begin{center}
\includegraphics[width=0.9\textwidth,height=11cm ,trim={0 0 0 2cm},clip]{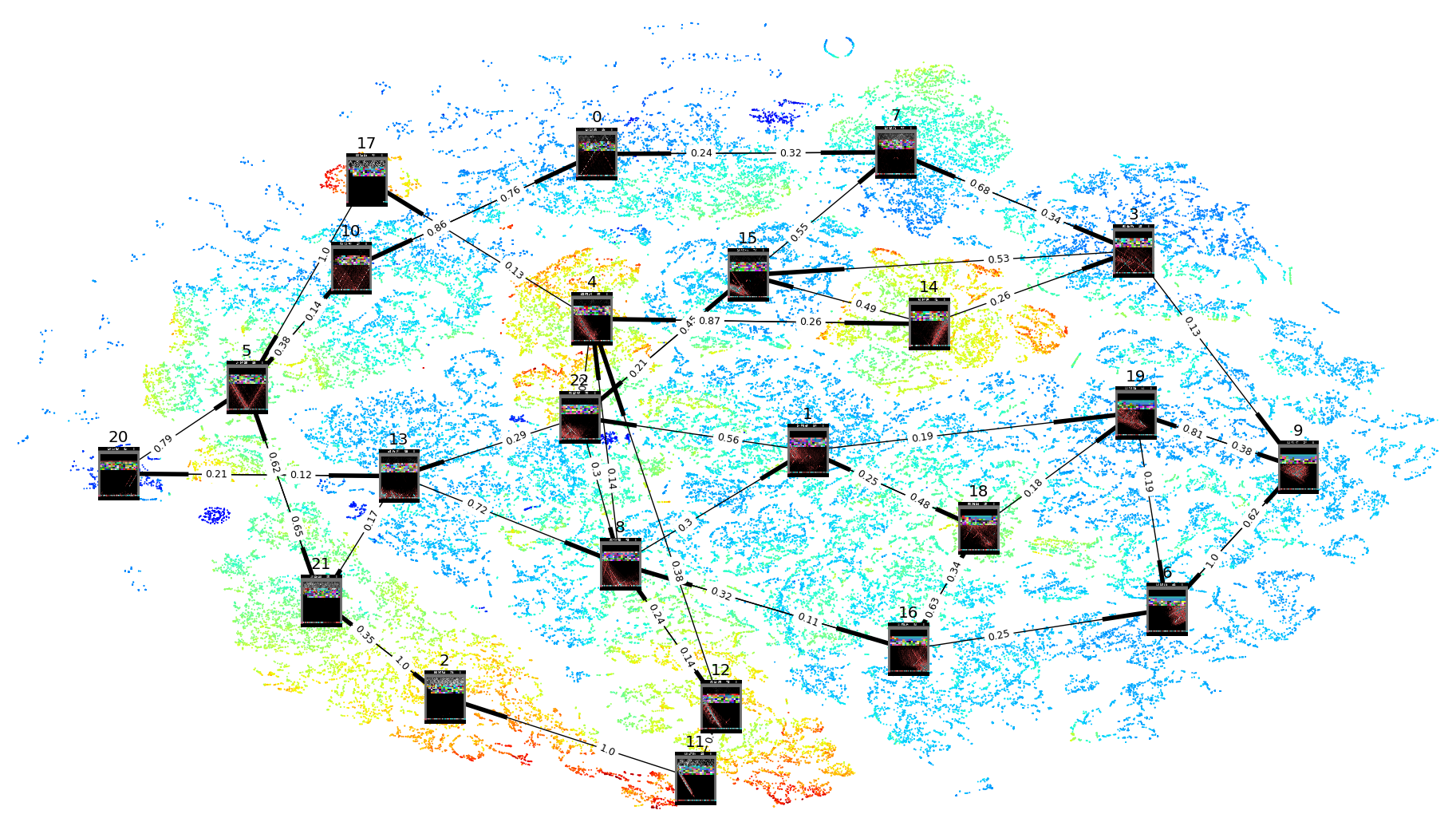} 
\caption{SAMDP visualization for Breakout. Each MDP state is represented on the map by its two t-SNE coordinates and colored by its value estimate (low values in blue and high in red). SAMDP states are visualize by their mean state (with frame pixels) at the mean t-SNE coordinate. An Edge between two SAMDP states represent a skill (bold side indicate terminal state), and the numbers above the edges correspond to the inferred SAMDP policy.}
\label{fig:Breakout}
\end{center}
\end{figure}

\textbf{SAMDP.} Figure~\ref{fig:Breakout} presents the resulted SAMDP for the game Breakout. First, we can observe that the algorithm managed to identify most of the policy hierarchy automatically and that the clusters are consistent with our manual clusters. We can also observe that the transitions between the clusters are sparse, which implies low entropy SAMDP. Finally, inspecting the mean image of each cluster reveals insights about the nature of the skills hiding within and uncovers the policy hierarchy.\\

\newpage
\subsection{Seaquest}

\begin{figure}
\begin{center}
\centerline{\includegraphics[trim=0cm 5cm 0cm 3cm,clip,width=16cm]{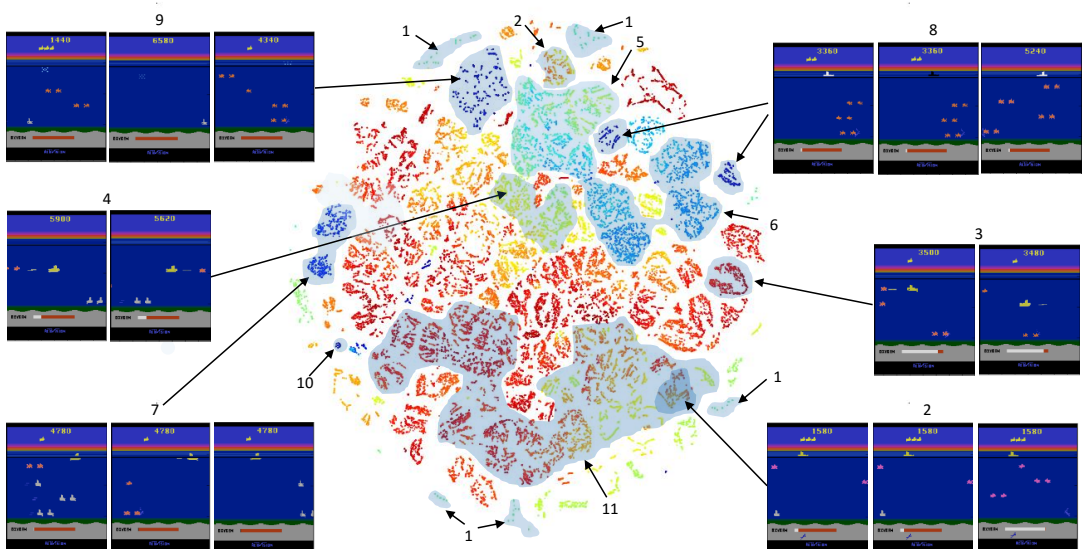}} 
\caption{Seaquest aggregated states on the t-SNE map, colored by value function estimate.}
\label{Seaquest_Aggregated}
\end{center}
\end{figure}
\textbf{Description.} In \href{https://archive.org/details/Seaquest_1983_Activision}{Seaquest}, the player's goal is to retrieve as many treasure-divers, while dodging and blasting enemy subs and killer sharks before the oxygen runs out. When the game begins, each enemy is worth 20 points and a rescued diver worth 50. Every time the agent surface with six divers, killing an enemy (rescuing a diver) is increased by 10 (50) points up to a maximum of 90 (1000). Moreover, the agent is awarded an extra bonus based on its remaining oxygen level. However, if it surfaces with less than six divers the oxygen fills up with no bonus, and if it surfaces with none it loses a life.\\

DQN's performance on Seaquest ($\sim$5k) is inferior to human experts ($\sim$100k). What makes Seaquest hard for DQN is that shooting enemies is rewarded immediately while rescuing divers is rewarded only once six divers are collected and rescued to sea level. Moreover, the bonus points for collecting 6 divers is diminished by reward clipping. This sparse and delayed reward signal requires much longer planning that is harder to learn.\\

\begin{figure}
\centering
    \begin{subfigure}

        \includegraphics[trim=2cm 11cm 4cm 2cm,clip,width=7cm]{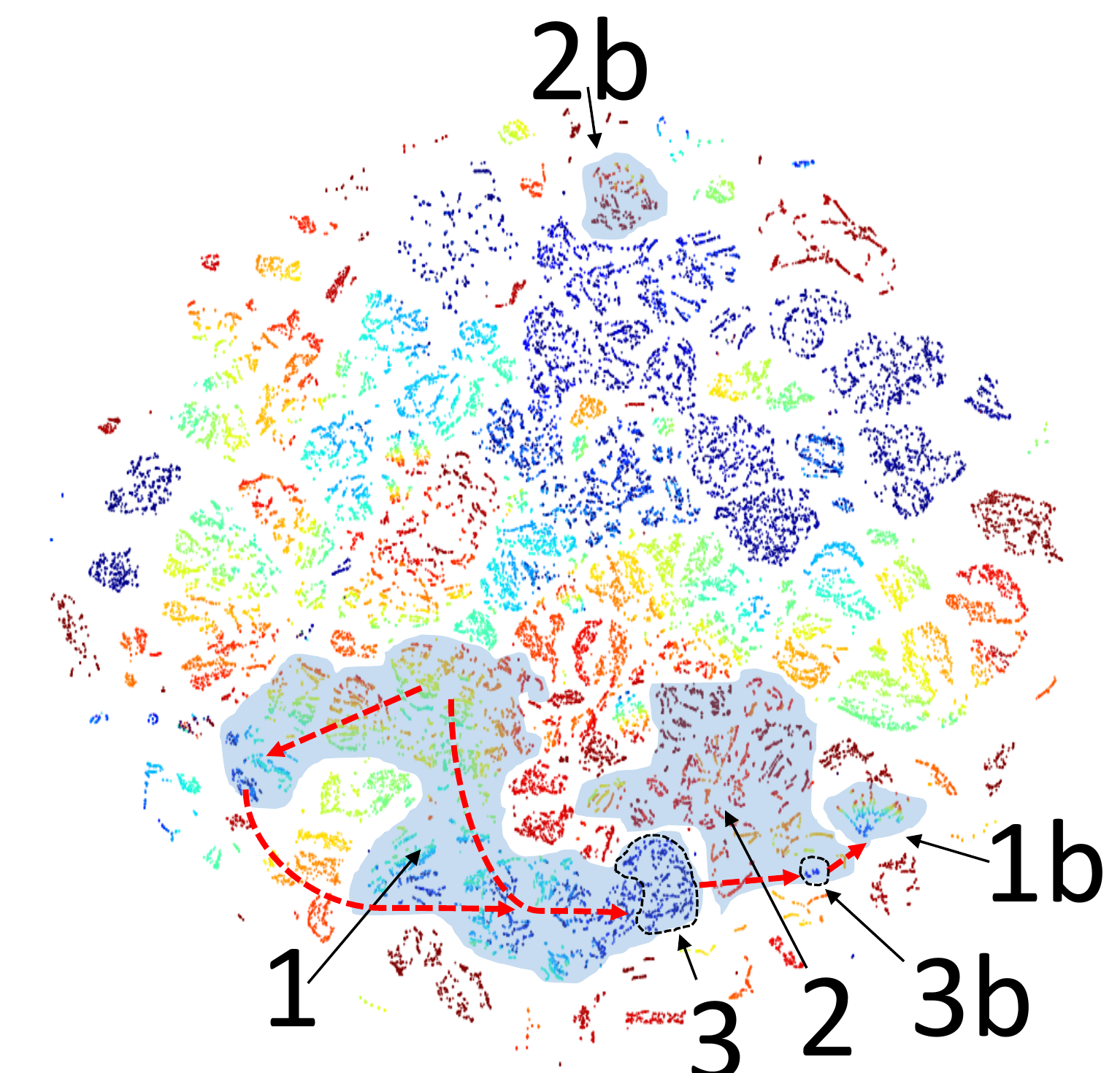}
    \end{subfigure}
    \begin{subfigure}
        
        \includegraphics[trim=0cm 5cm 0cm 5cm,clip,width=7cm]{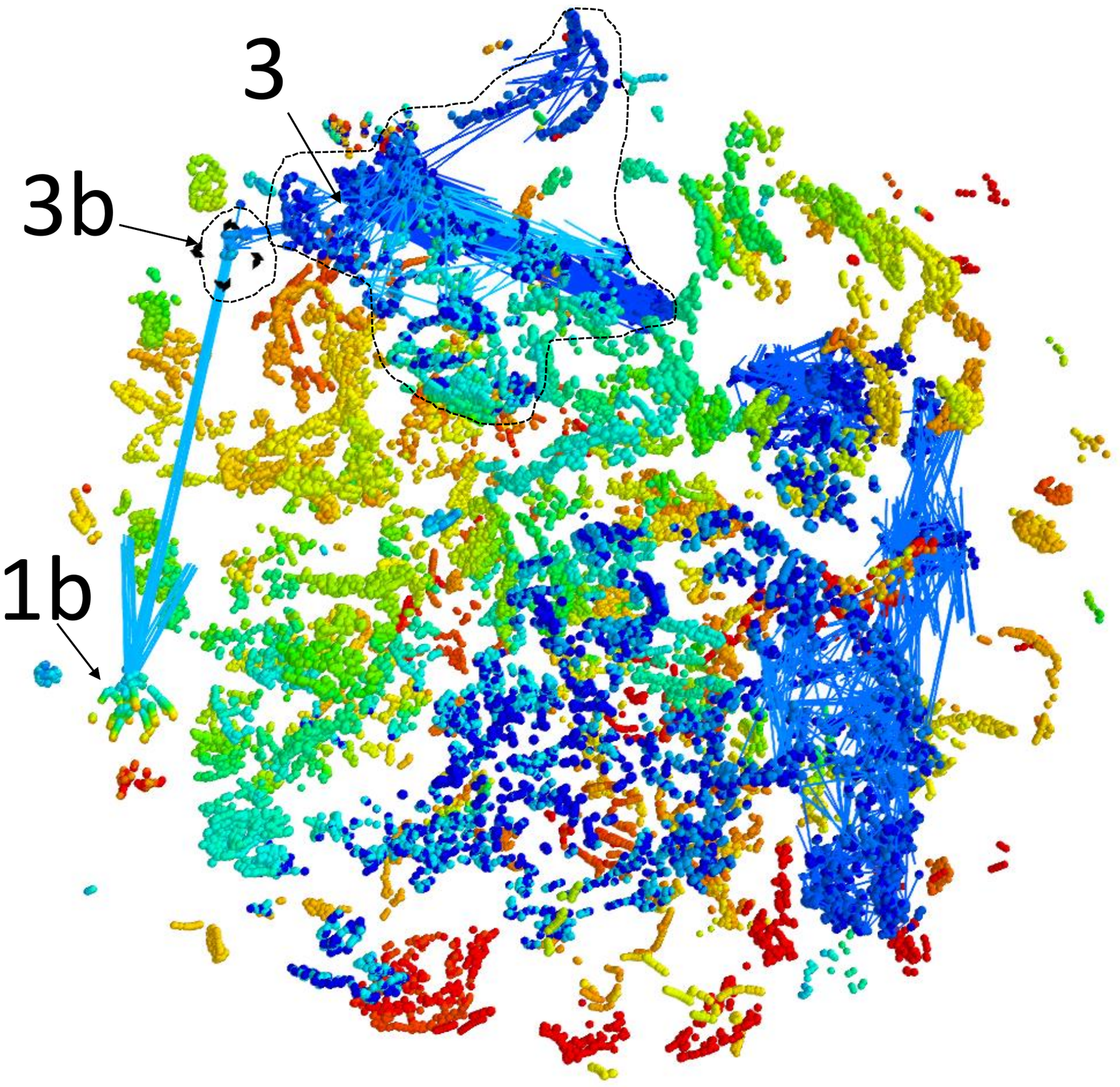}
    \end{subfigure}
\caption{Seaquest refuel option. \textbf{Left:} Option path above the t-SNE map colored by the amount of remaining oxygen. Shaded blue mark the clusters with a collected diver, and red arrows mark the direction of progress. \textbf{Right:} 3d t-SNE with colored arrows representing transitions. All transitions from cluster 3 are leading to cluster 3b in order to reach the refuel cluster (1b), thus indicating a clear option termination structure.}
\label{seaquestOption}
\end{figure}

\textbf{Hand-crafted features.} We extract features for player position, direction of movement (ascent/descent), oxygen level, number of enemies, number of available divers to collect, number of available divers to use, and number of lives.\\

\textbf{Analysis.} Figure~\ref{Seaquest_Aggregated} shows the t-SNE map divided into different clusters. We notice two main partitions of the t-SNE clusters: (a) by oxygen level (low: clusters 4-10, high: cluster 3 and other unmarked areas), and (b) by the amount of collected divers (clusters 2 and 11 represent having a diver). We also identified other partitions between clusters such as refuelling clusters (1: pre-episode and 2: in-episode), various termination clusters (8: agent appears in black and white, 9: agent's figure is replaced with drops, 10: agent's figure disappears) and low oxygen clusters characterized by flickering in the oxygen bar (4 and 7). 

\begin{figure}
\begin{center}
\centerline{\includegraphics[width=\columnwidth,trim=3cm 20cm 3cm 3cm,clip]{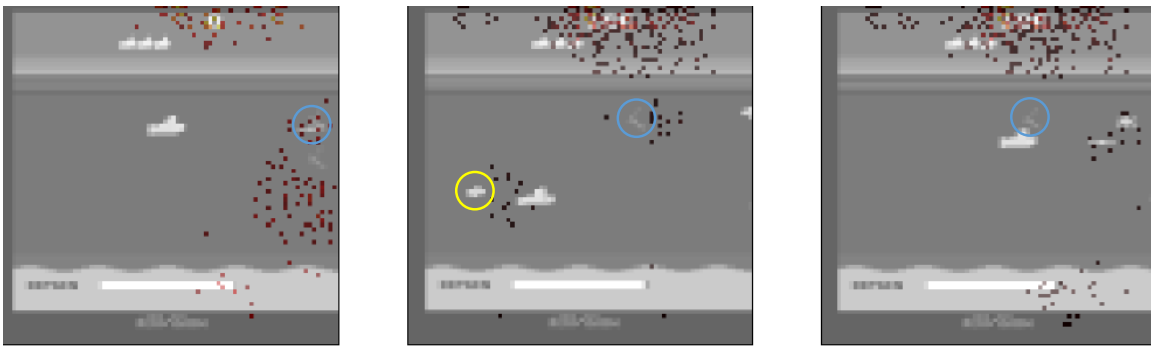}}
\caption{Salincy maps of states with available diver. \textbf{Left:} A diver is noticed in the saliency map but misunderstood as an enemy and being shot at. \textbf{Center:} Both diver and enemy are noticed by the network. \textbf{Right:} The diver is unnoticed by the network.}
\label{SeaquestSaliencyDiver}
\end{center}
\end{figure} 

While the agent distinguishes states with a collected diver, Figure~\ref{SeaquestSaliencyDiver} implies that the agent did not understand the concept of collecting a diver and sometimes treats it as an enemy. Moreover, we see that the clusters share a similar value estimate that is highly correlated with the amount of remaining oxygen and the number of present enemies. However, states with an available or collected diver do not raise the value estimates nor do states that are close to refueling. Moreover, the agent never explores the bottom of the screen, nor collects more than two divers.\\ 

As a result, the agent's policy is to kill as many enemies as possible while avoiding being shot. If it hasn't collected a diver, the agent follows a sub-optimal policy and ascends near to sea level as the oxygen decreases. There, it continues to shoot at enemies but not collecting divers. However, it also learned not to surface entirely without a diver.

If the agent collects a diver it moves to the blue shaded clusters (all clusters in this paragraph refer to Figure~\ref{seaquestOption}), where we identify a refuel option. We noticed that the option can be initiated from two different zones based on the oxygen level but has a singular termination cluster (3b). If the diver is taken while having a high level of oxygen, then it enters the option at the northern (red) part of cluster 1. Otherwise, it will enter a point further along the direction of the option (red arrows). In cluster 1, the agent keeps following the normal shooting policy. Eventually, the agent reaches a critical level of oxygen (cluster 3) and ascends to sea level. From there the agent jumps to the fueling cluster (area 1b). The fueling cluster is identified by its rainbow appearance because the level of oxygen is increasing rapidly. However, the refuel option was not learned perfectly. Area 2 is another refuel cluster, there, the agent does not exploit its oxygen before ascending to get air (area 2b).

\newpage
\subsection{Pacman}

\begin{figure}
\begin{center}
\centerline{\includegraphics[trim=5cm 8.5cm 2cm 6cm,clip,width=0.6\textwidth]{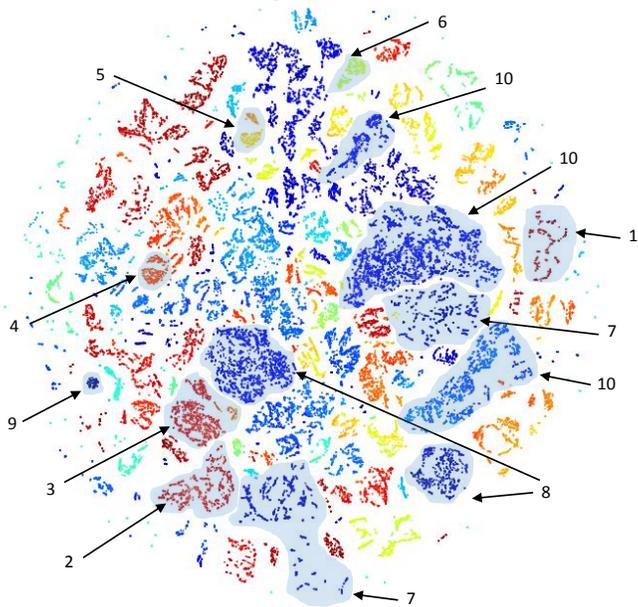}} 
\caption{Pacman aggregated states on the t-SNE map colored by value function estimates.}
\label{Pacman}
\end{center}
\end{figure}

\textbf{Description.}
In \href{https://archive.org/details/manuals-handheld-games-Coleco-PacMan}{Pacman}, an agent navigates in a maze while being chased by two ghosts. The agent is positively rewarded (+1) for collecting bricks. An episode ends when a predator catches the agent, or when the agent collects all bricks. There are also 4 bonus bricks, one at each corner of the maze. The bonus bricks provide a larger reward (+5), and more importantly, they make the ghosts vulnerable for a short period of time, during which they cannot kill the agent. Occasionally, a bonus box appears for a short time providing high reward (+100) if collected.\\

\textbf{Hand-crafted features.} We extract features for the player position, direction of movement, number of left bricks to eat, minimal distance (L1) between the player and the predators, number of lives, “ghost mode” that indicates that the predators are vulnerable, and “bonus box” feature that indicate when the highly valued box appears.\\

\begin{figure}
\vskip 0.2in
\begin{center}
\centerline{\includegraphics[width=8cm,trim=1cm 4.5cm .5cm 1.5cm,clip]{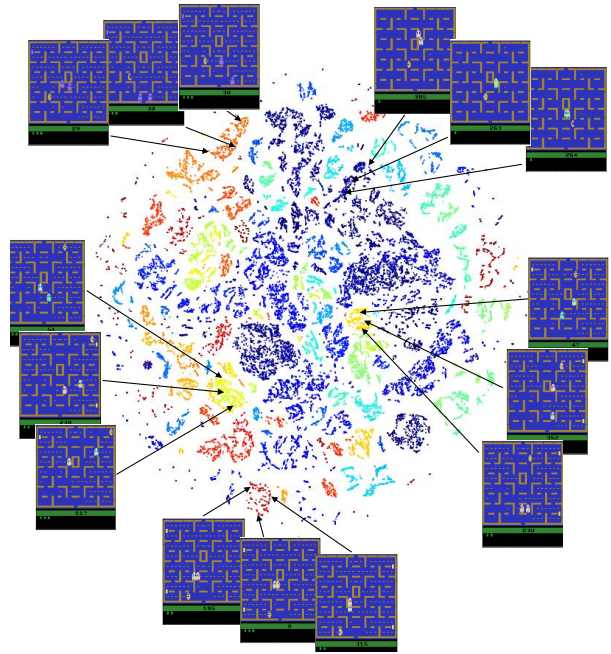}}
\caption{t-SNE for Pacman colored by the number of left bricks with state examples from each cluster.}
\label{Pacman_bricks}
\end{center}
\vskip -0.2in
\end{figure}

\textbf{Analysis.}
Figure~\ref{Pacman} shows the t-SNE colored by value function, while Figure~\ref{Pacman_bricks} shows the t-SNE colored by the number of left bricks with examples of states from each cluster. We can see that the clusters are well partitioned by the number of remaining bricks and value estimates. Moreover, examining the states in each cluster (Figure~\ref{Pacman_bricks}) we see that the clusters share specific bricks pattern and agent location.

\begin{figure}
\begin{center}
\centerline{\includegraphics[width=7cm,trim=2cm 15.5cm 4cm 2.5cm,clip]{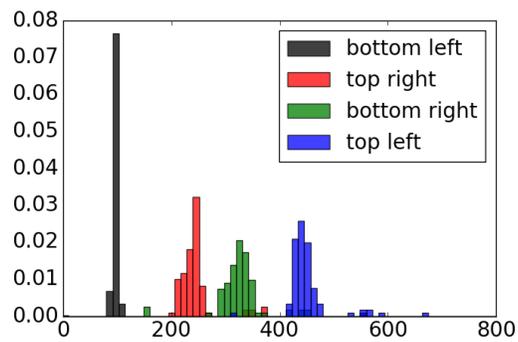}}
\caption{Histogram showing the time passed until each bonus brick is collected.}
\label{Pacman_bonus_bricks}
\end{center}
\end{figure} 

From Figure~\ref{Pacman_bonus_bricks} we also learn that the agent collects the bonus bricks at very specific times and order. Thus, we conclude that the agent has learned a location-based policy that is focused on collecting the bonus bricks (similar to maze problems) while avoiding the ghosts.

\begin{figure}[h]
\begin{center}
\includegraphics[trim=2cm 1cm 2cm 1cm,clip,width=\textwidth]{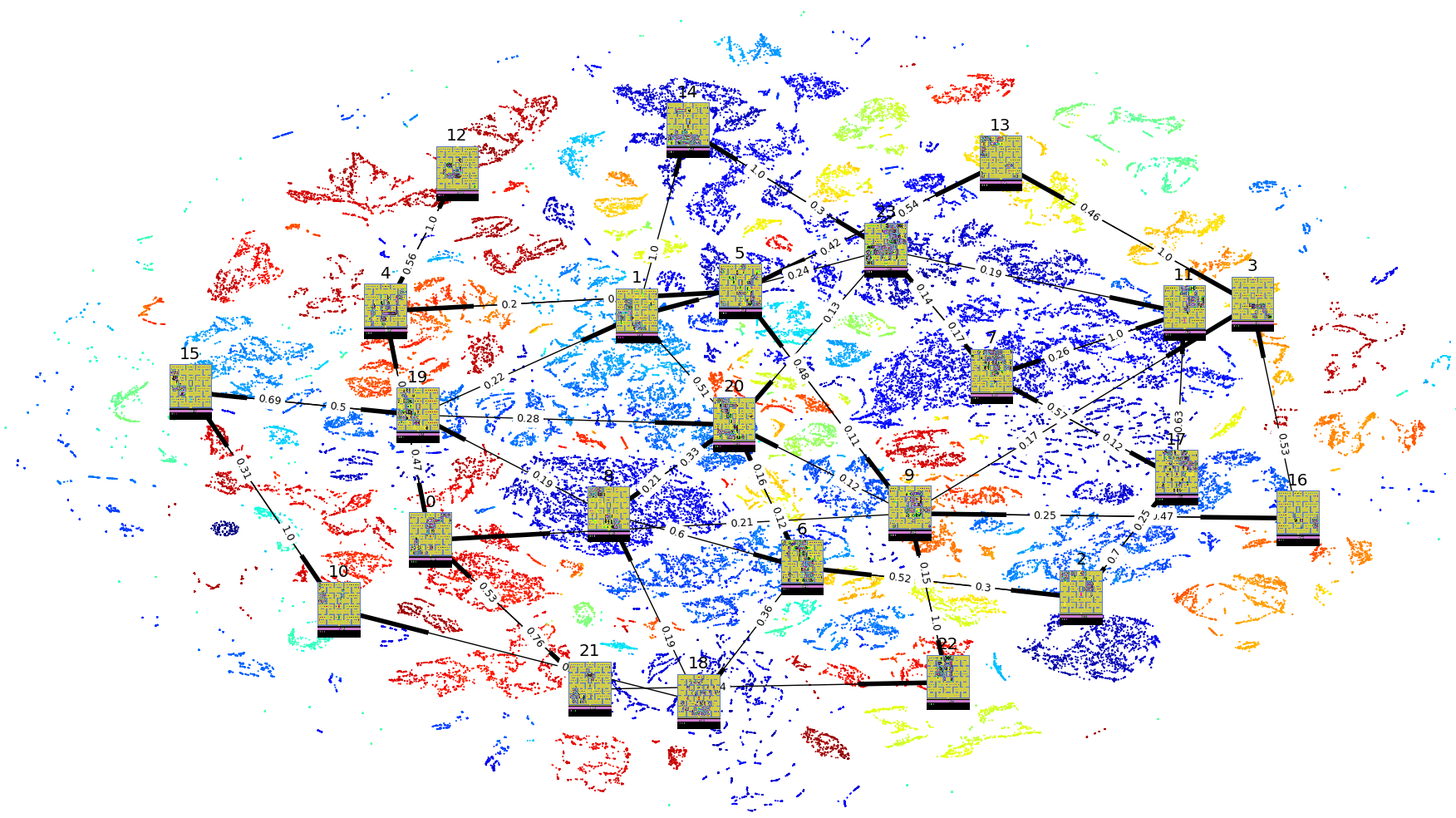} 
\caption{SAMDP visualization for Pacman.}
\label{fig:Pacman}
\end{center}
\end{figure} 

The agent starts at cluster 1, heading for the bottom-left bonus brick and collecting it in cluster 2. Once collected, it moves to cluster 3 where it is heading to the top-right bonus brick and collects it in cluster 4. The agent is then moving to clusters 5 and 6 where it is taking the bottom-right and top-left bonus bricks respectively. We also identified cluster 7 and 9 as termination clusters and cluster 8 as a hiding cluster in which the agent hides from the ghosts in the top-left corner. The effects of reward-clipping can be clearly seen in the case of the bonus box. Cluster 10 comprises of states with a visible bonus box. However, these states are assigned with a lower value.\\

\textbf{SAMDP.} Figure~\ref{fig:Pacman} presents the resulted SAMDP for the game Pacman. The sparsity of the transitions between the clusters indicates that the agent is spatio-temporally located, meaning that it spends long times in defined areas of the state space as suggested by Figure~\ref{Pacman_bonus_bricks}.

\newpage
\subsection{Enviroment modeling}
\begin{figure}[h]
\begin{center}
\centerline{\includegraphics[width=\columnwidth,trim=0.5cm 2cm 2cm 1.5cm,clip]{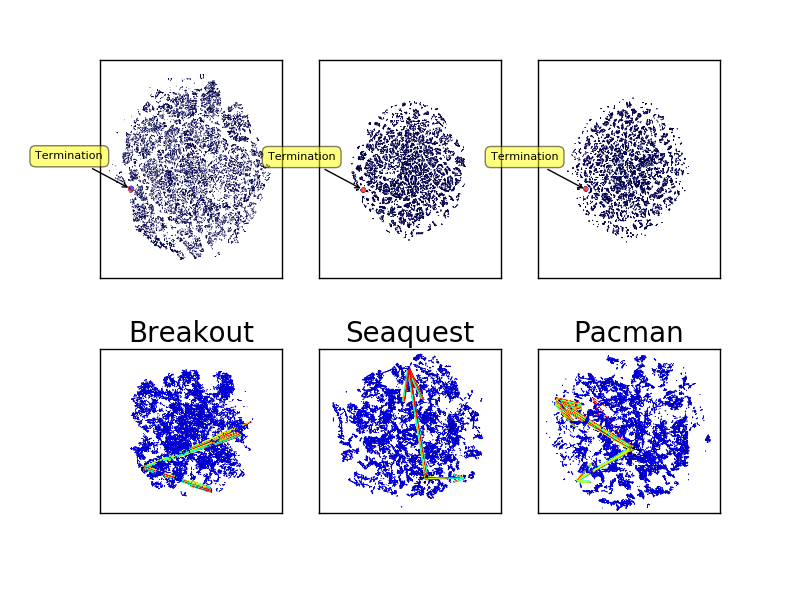}}
\caption{Terminal and initial states. \textbf{Top:} All terminal states are mapped into a singular zone(red). \textbf{Bottom:} Initial states are mapped into singular zones (pointed by colored arrows from the terminal zone) above the 3d t-SNE dynamics representation.}
\label{Terminal-initial}
\end{center}
\end{figure}

The DQN algorithm requires specific treatment for initial (padded with zero frames) and terminal (receive target value of zero) states, however, it is not clear how to check if this treatment works well. Therefore we show t-SNE maps for different games with a focus on termination and initial states in Figure~\ref{Terminal-initial}. We can see that all terminal states are mapped successfully into a singular zone, however, initial states are also mapped to singular zones and assigned with wrong value predictions. Following these observations, we suggest to investigate different ways to model initial states, i.e., replicating a frame instead of feeding zeros and test it.

\subsection{Score pixels} \label{score_section}
\begin{figure}[h]
\begin{center}
\centerline{\includegraphics[width=\columnwidth,trim=0cm 0cm 0cm 0cm,clip]{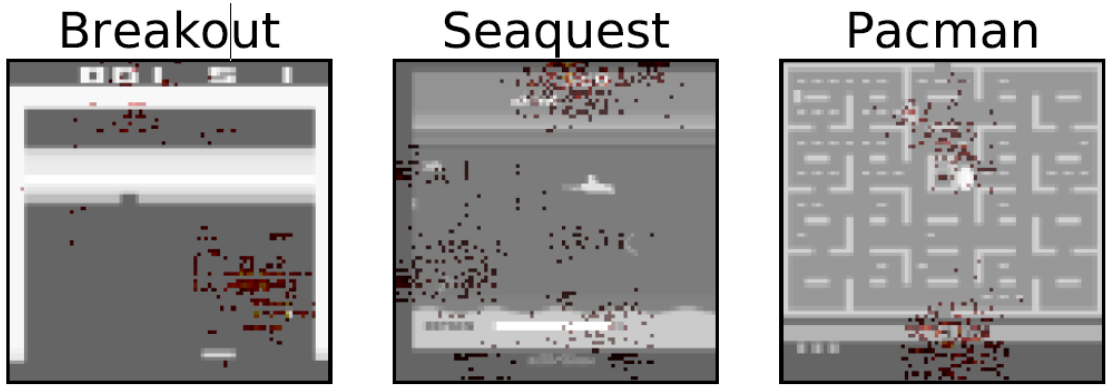}}
\caption{Saliency maps of score pixels. The input state is presented in gray scale while the value input gradients are displayed above it in red.}
\label{SaliencyScore}
\end{center}
\end{figure}

Some Atari2600 games include multiple repetitions of the same game. Once the agent finished the first screen it is presented with another one, distinguished only by the score that was accumulated in the first screen. Therefore, an agent might encounter problems with generalizing to the new screen if it over-fits the score pixels. In Breakout, for example, the current state of the art architectures achieves a game score of around 450 points while the maximal available points are 896 suggesting that the agent is somehow not learning to generalize for the new level. We investigated the effect that score pixels have on the network predictions. Figure~\ref{SaliencyScore} shows the saliency maps of different games supporting our claim that DQN is basing its estimates using these pixels. We suggest to further investigate this, for example, we suggest to train an agent that does not receive those pixels as input. 

\newpage
\subsection{SAMDPs}

\begin{figure}
\centering
\includegraphics[width=15cm, height=6.5cm, trim=2cm 0cm 0cm 1cm,clip]{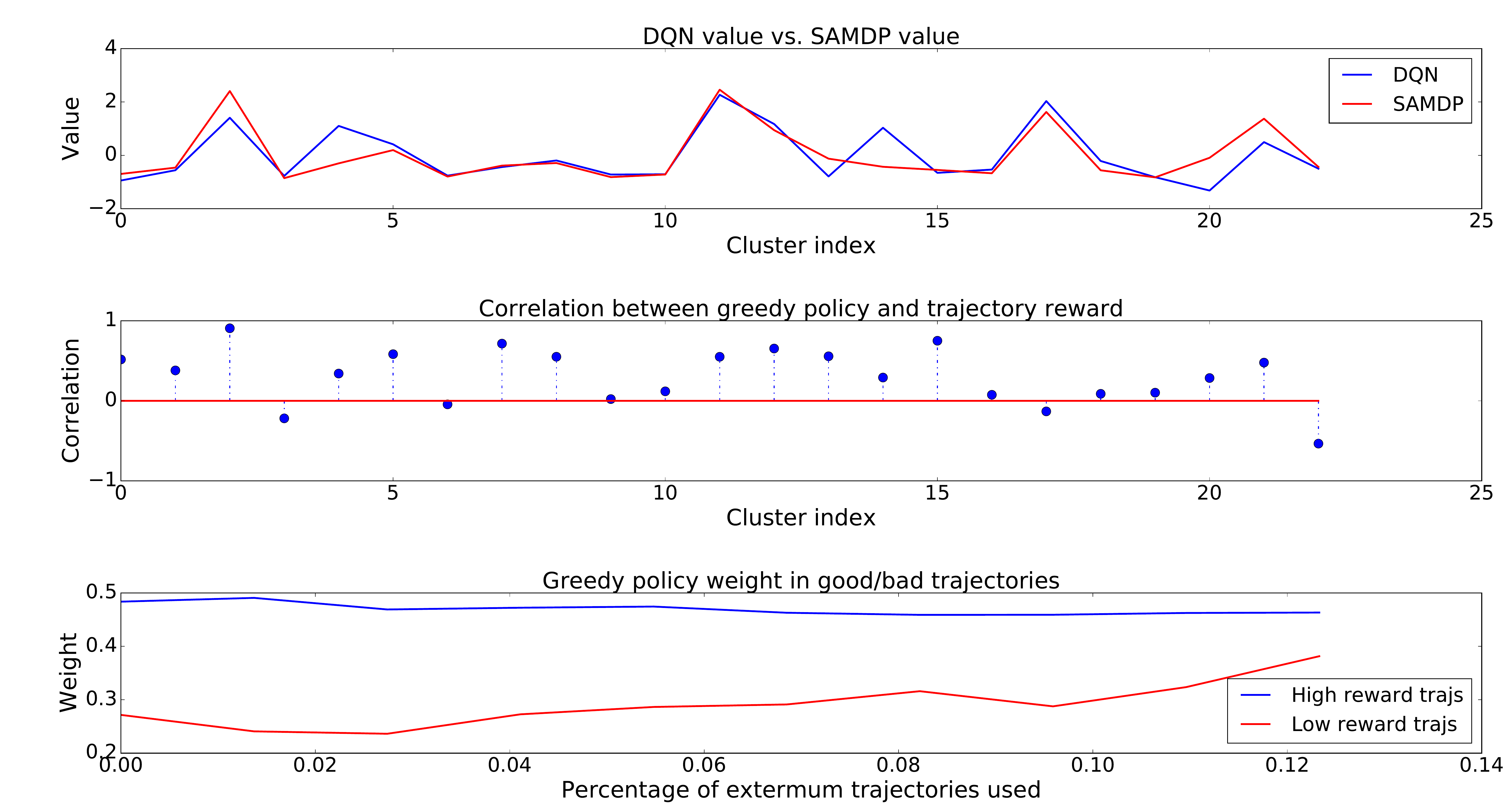}
\caption{\textbf{Model Evaluation.} \textbf{Top:} VMSE. \textbf{Center:} greedy policy correlation with trajectory reward. \textbf{Bottom:} top (blue), least (red) rewarded trajectories.}
\label{fig:model_consis}
\end{figure}

\textbf{Model Evaluation.} We perform three simulations to evaluate the SAMDP model. First, we measure the VMSE criteria, as defined in Equation \ref{eq:vmse}. Since the true value function $v$ is unknown, we approximate it by averaging the DQN value estimates in each cluster: $${v^{DQN}(c_j)}=\frac{1}{|C_j|}\sum_{i: s_i \in c_j}v^{DQN}(s_i),$$ and evaluate $V_{SAMDP}$ as defined in Section~\ref{sec:samdp}. Examining Figure~\ref{fig:model_consis}, top, we observe that DQN values and the SAMDP values are very similar, indicating that the SAMDP model fits the data well.\\

Second, we evaluate the greedy policy with respect to the SAMDP value function by: $\pi_{greedy}(c_i) = \underset{j}{\mbox{argmax}} \{ R_{\sigma_{i,j}}+\gamma^{k_{\sigma_{i,j}}}v_{SAMDP}(c_j) \}$. We measure the correlation between the greedy policy decisions and the trajectory reward. For a given trajectory $j$ we measure $P_i^j:$ the empirical distribution of choosing the greedy policy at state $c_i$ and the cumulative reward  $R^j$. Finally, we present the correlation between these two measures in each state: $corr_i = corr(P_i^j,R^j)$ in (Figure~\ref{fig:model_consis}, center). Positive correlation indicates that following the greedy policy leads to high reward. Indeed for most of the states we observe positive correlation, supporting the consistency of the model.\\

The third evaluation is close in spirit to the second one. We partition the data to train and test. We evaluate the greedy policy based on the train set and create two transition matrices $T^+,T^-$ using k top and bottom rewarded trajectories respectively from the test set. We measure the correlation of the greedy policy $T^G$ with each of the transition matrices for different values of k (Figure~\ref{fig:model_consis} bottom). As clearly seen, the correlation of the greedy policy and the top trajectories is higher than the correlation with the bottom trajectories.\\

\textbf{Eject Button:} The motivation for this experiment stems from the idea of shared autonomy \citep{icra11a}. There are domains where errors are not permitted and performance must be as high as possible. The idea of shared autonomy is to allow an operator to intervene in the decision loop in critical times. For example, it is known that in $20\%$ of commercial flights, the auto-pilot returns the control to the human pilots. In the following experiment, we show how the SAMDP model can help to identify where agent's behavior deteriorates. \textbf{Setup.} \textit{(a)} Evaluate a DQN agent, create a trajectory data set, and evaluate the features for each state (stage 0). \textit{(b)} Divide the data between train (100 trajectories) and test (60) and build an SAMDP model (stages 1-4) on the train data. \textit{(c)} Split the train data to k top and bottom rewarded trajectories $T^+,T^-$ and re-infer the model parameters separately for each (stage 3). \textit{(d)} Project the test data on the SAMDP model (each state is mapped to the nearest SAMDP state). \textit{(e)} Once it becomes more likely that the agent transitions are coming from $T^-$ rather then from $T^+$, we \textbf{eject} (inspired by option interruption \citep{sutton1999between}). \textit{(f)} We average the trajectory reward on \textit{(i)} the entire test set, and \textit{(ii)} the un-ejected trajectories subset. We measure $36\%$, $20\%$, and $4.7\%$ performance gain for Breakout Seaquest and Pacman, respectively. The eject experiment indicates that the SAMDP model can be used to robustify a given DQN policy by identifying when the agent is not going to perform well and return control to a human operator or some other AI agent.

\section{Conclusions}
In this work, we showed that the features learned by DQN map the state space to different sub-manifolds, in each, different features are present. By analyzing the dynamics in these clusters and between them we were able to identify hierarchical structures. In particular, we were able to identify options with defined initial and termination rules. State aggregation gives the agent the ability to learn specific policies for the different regions, thus giving an explanation to the success of DRL agents. The ability to automatically detect sub-goals and skills can also be used in the future as a component of an hierarchical DRL algorithm \citep{tessler2016deep,kulkarni2016hierarchical}. Another possibility is to use the aggregation mechanism in order to allocate learning resources between clusters better, for example by using a prioritized experience replay \citep{schaul2015prioritized}. 

Similar to Neuro-Science, where reverse engineering methods like fMRI reveal structure in brain activity, we demonstrated how to describe the agent's policy with simple logic rules by processing the network's neural activity. We believe that interpretation of policies learned by DRL agents is of particular importance by itself.  First, it can help in the debugging process by giving the designer qualitative understanding of the learned policies. Second, there is a growing interest in applying DRL solutions to a real-world problem such as autonomous driving and medicine. We believe that before we can reach that goal we will have to gain greater confidence on what the agent is learning. Lastly, understanding what is learned by DRL agents, can help designers develop better algorithms by suggesting solutions that address policy downsides.\\

To conclude, when a deep network is not performing well it is hard to understand the cause and even harder to find ways to improve it. Particularly in DRL, we lack the tools needed to analyze what an agent has learned and therefore left with black box testing. In this paper we showed how to gray the black box: understand better why DQNs work well in practice, and suggested a methodology to interpret the learned policies.

In the Eject experiment, we showed how SAMDP model can help to improve the policy at hand without the need to re-train it. It would be even more interesting to use the SAMDP model to improve the training phase itself.

\section*{Acknowledgement}
This research was supported in part by the European Community’s Seventh Framework Programme (FP7/2007-2013) under grant agreement 306638 (SUPREL) and the Intel Collaborative Research Institute for Computational Intelligence (ICRI-CI).

\clearpage

\bibliographystyle{icml2016}
\bibliography{paper_bib}
\newpage

\section{Appendix}

\begin{figure}[h]
\begin{center}
\centerline{\includegraphics[trim=0cm 5cm 0cm 3cm,clip,width=15cm]{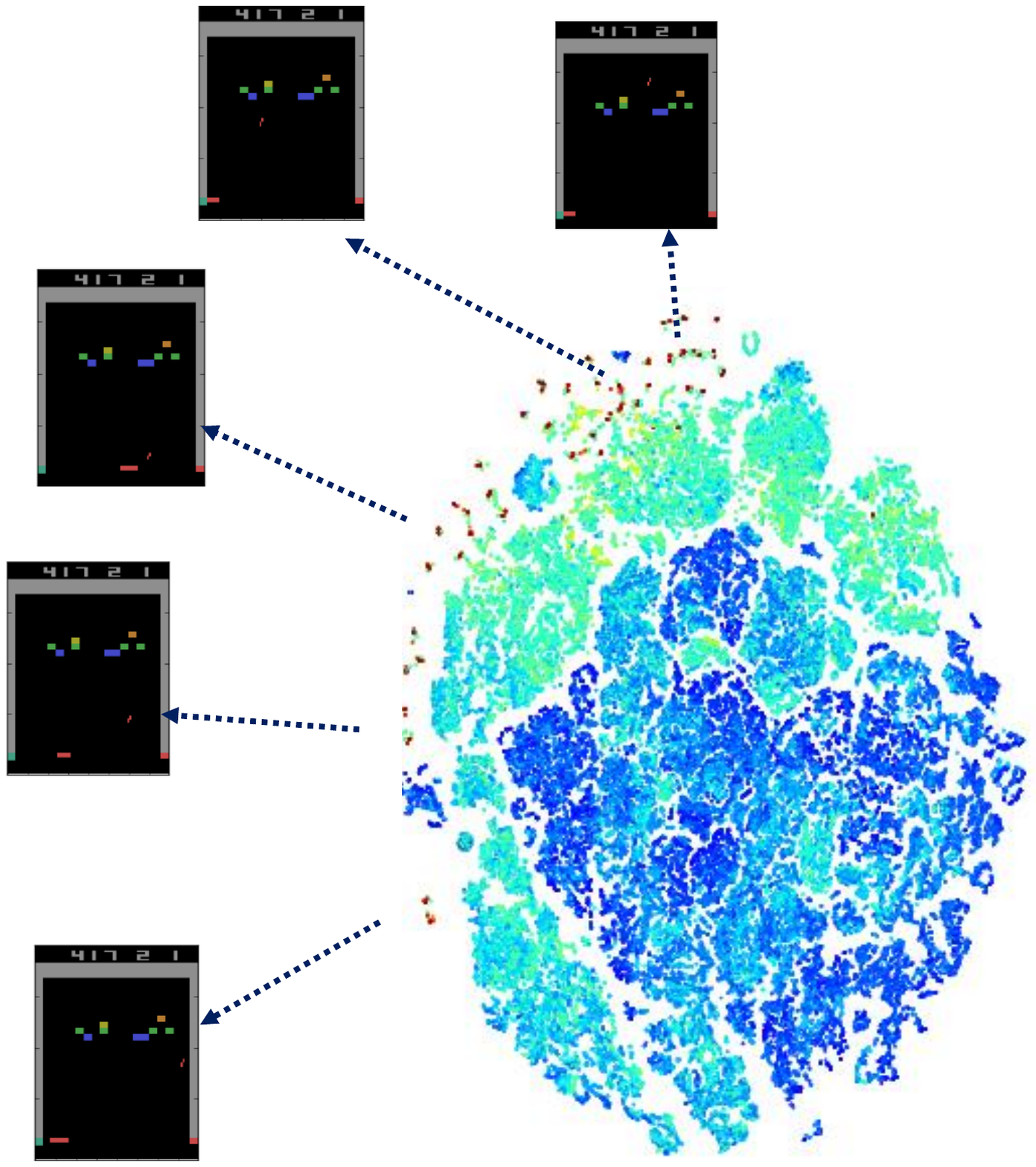}} 
\caption{t-SNE for Breakout colored by time, with examples of states that are repeated by the agent multiple times.}
\label{BreakoutTime}
\end{center}
\end{figure} 

\begin{figure}
\begin{center}
\centerline{\includegraphics[trim=0cm 5cm 0cm 3cm,clip,width=16cm]{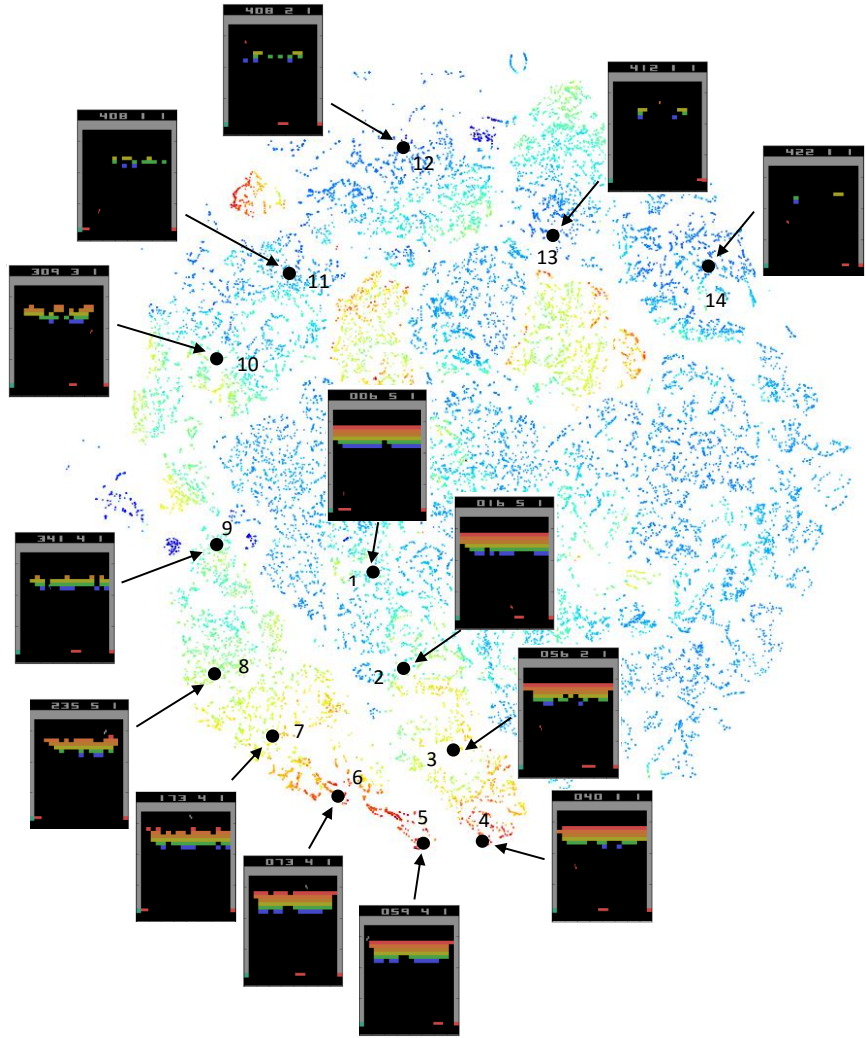}} 
\caption{Breakout t-sne colored by value. Example of trajectory along the t-sne map.}
\label{BreakoutOption}
\end{center}
\end{figure}

\begin{figure}
\vskip 0.2in
\begin{center}
\centerline{\includegraphics[trim=0cm 15cm 0cm 3cm,clip,width=12cm]{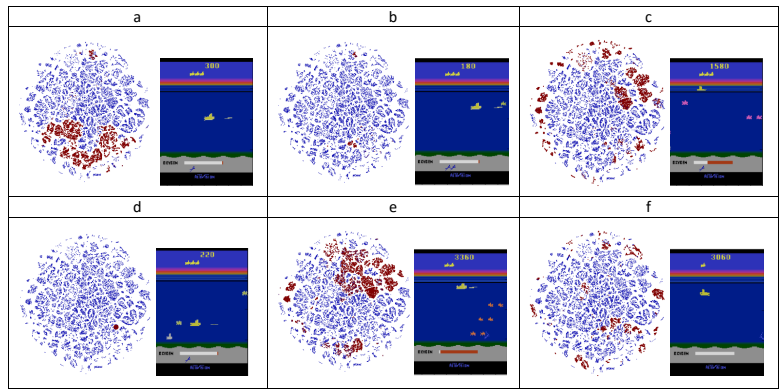}} 
\caption{Seaquest states partitioned on the t-SNE map by different measure. a: Number of taken divers = 1. b: Number of taken divers = 2. c: Agent is next to sea level (filter by vertical position $<$ 60). d: Agent is never visiting the lower 1/3 part of the frame. We mark the state of maximal diving depth (filter by vertical position $>$ 128). e: Oxygen level $<$ 0.05. f: Oxygen level $>$ 0.98.}
\label{SeaquestFiltering}
\end{center}
\vskip -0.2in
\end{figure} 
%\begin{figure}
%\vskip 0.2in
%\begin{center}
%\centerline{\includegraphics[trim=0cm 0cm 0cm 0cm,clip,width=16cm]{../figs/Pacman_tSNE}} 
%\caption{Pacman t-SNE with example states.}
%\label{Pacman_tSNE}
%\end{center}
%\vskip -0.2in
%\end{figure} 
\end{document}